\documentclass{article}

\usepackage[final]{neurips_2025}

\usepackage[utf8]{inputenc} 
\usepackage[T1]{fontenc}    
\usepackage{hyperref}       
\usepackage{url}            
\usepackage{booktabs}       
\usepackage{amsfonts}       
\usepackage{nicefrac}       
\usepackage{microtype}      
\usepackage{xcolor}         

\usepackage{times}
\usepackage{latexsym}
\usepackage{graphicx}
\usepackage{amsmath}
\usepackage{amssymb}
\usepackage{mathtools}
\usepackage{amsthm}
\usepackage{array}
\usepackage{longtable}
\usepackage[most]{tcolorbox}
\usepackage{bibentry}
\usepackage{bm} 
\usepackage{tabularx}
\usepackage{multirow} 
\usepackage{algorithm}
\usepackage{algorithmic}
\usepackage{arydshln}
\usepackage{appendix}
\usepackage{wrapfig}   
\usepackage{bbm}

\makeatletter
\def\blankfootnote{\xdef\@thefnmark{}\@footnotetext}
\makeatother

\definecolor{royalblue(web)}{rgb}{0.25, 0.41, 0.88}
\definecolor{blue-violet}{rgb}{0.54, 0.17, 0.89}
\definecolor{brightmaroon}{rgb}{0.76, 0.13, 0.28}
\definecolor{darkmagenta}{rgb}{0.55, 0.0, 0.55}
\definecolor{bleudefrance}{rgb}{0.19, 0.55, 0.91}
\definecolor{palatinateblue}{rgb}{0.15, 0.23, 0.89}
\definecolor{royalblue(web)}{rgb}{0.25, 0.41, 0.88}
\definecolor{whitesmoke}{rgb}{0.96, 0.96, 0.96}
\definecolor{thulianpink}{rgb}{0.87, 0.44, 0.63}
\definecolor{amber(sae/ece)}{rgb}{1.0, 0.49, 0.0}
\definecolor{darkblue}{rgb}{0.0, 0.0, 0.55}
\definecolor{alizarin}{rgb}{0.82, 0.1, 0.26}
\definecolor{asparagus}{rgb}{0.53, 0.66, 0.42}
\definecolor{darkspringgreen}{rgb}{0.09, 0.45, 0.27}
\definecolor{columbiablue}{rgb}{0.61, 0.87, 1.0}
\definecolor{wildblueyonder}{rgb}{0.64, 0.68, 0.82}
\definecolor{trolleygrey}{rgb}{0.5, 0.5, 0.5}
\definecolor{paleaqua}{rgb}{0.74, 0.83, 0.9}
\definecolor{bubblegum}{rgb}{0.99, 0.76, 0.8}
\definecolor{coralred}{rgb}{1.0, 0.25, 0.25}
\definecolor{green(ryb)}{rgb}{0.4, 0.69, 0.2}
\definecolor{flame}{rgb}{0.89, 0.35, 0.13}
\definecolor{bittersweet}{rgb}{1.0, 0.44, 0.37}
\definecolor{darksalmon}{rgb}{0.91, 0.59, 0.48}
\definecolor{emerald}{rgb}{0.31, 0.78, 0.47}
\definecolor{green(pigment)}{rgb}{0.0, 0.65, 0.31}
\definecolor{codegreen}{rgb}{0,0.6,0}
\definecolor{codegray}{rgb}{0.5,0.5,0.5}
\definecolor{codepurple}{rgb}{0.58,0,0.82}
\definecolor{backcolour}{rgb}{0.96,0.96,0.94}
\definecolor{bluegray}{rgb}{0.3, 0.38, 0.47}
\definecolor{whitesmoke}{rgb}{0.96, 0.96, 0.96}
\definecolor{codegreen}{rgb}{0,0.6,0}
\definecolor{codegray}{rgb}{0.5,0.5,0.5}
\definecolor{codepurple}{rgb}{0.58,0,0.82}
\definecolor{backcolour}{rgb}{0.96,0.96,0.94}

\tcbuselibrary{breakable}
\usepackage{listings}
\lstdefinestyle{mystyle}{
  basicstyle=\scriptsize\ttfamily,
  frame=single, 
  columns=fixed, 
}

\lstdefinestyle{newstyle}{
  basicstyle=\footnotesize\ttfamily\color{codegreen},
  backgroundcolor=\color{backcolour},
  frame=shadowbox, 
  rulecolor=\color{red},
  frameround=tttt, 
  keywordstyle=\color{magenta},
  commentstyle=\color{green},
  stringstyle=\color{red},
  showstringspaces=false,
  numbers=left,
  numberstyle=\tiny\color{gray},
  breaklines=true
}

\usepackage{color}

\newtheorem{theorem}{Theorem}

\newtheorem{proposition}{Proposition}
\newtheorem{definition}{Definition}

\newcommand{\ours}{{\fontfamily{qpl}\selectfont ReDit}}
\newcommand{\xbf}{{\mathbf x}}
\newcommand{\ybf}{{\mathbf y}}
\newcommand{\smax}{\mathrm{softmax}}
\newcommand{\nn}{f}
\newcommand{\X}{{\mathcal X}}
\newcommand{\EE}{\mathop{\mathbb E}}
\newcommand{\datasetpg}{S}
\newcommand{\gtreward}{r_{\mathrm{G}}}
\newcommand{\OO}{{\mathcal O}}

\newcommand{\1}{\mathbf{1}}

\usepackage{amsmath,amsfonts,bm}

\def\1{\bm{1}}

\DeclareMathAlphabet{\mathsfit}{\encodingdefault}{\sfdefault}{m}{sl}
\SetMathAlphabet{\mathsfit}{bold}{\encodingdefault}{\sfdefault}{bx}{n}

\newcommand{\var}{\mathrm{var}}

\title{\textbf{\ours{}}: Reward Dithering for Improved  LLM Policy Optimization}

\author{
  Chenxing Wei$^{\dagger \S \natural}$, Jiarui Yu$^{\circ}$, Ying He$^{ \dagger}$, Hande Dong$^{\circ}$, Yao Shu$^{ \wr}$\thanks{corresponding author. $^\natural$ Work done during an internship at Tencent.} , F. Richard Yu$^{ \ddagger}$\\
$^{\S}$Guangdong Lab of AI and Digital Economy (SZ), China\\
$^\dagger$College of Computer Science and Software Engineering, Shenzhen University, China\\
$^{\wr}$Hong Kong University of Science and Technology (Guangzhou), China \\
$^\circ$ Tencent, Shenzhen, China \\
$^{\ddagger}$School of Information Technology, Carleton University, Canada \\
\texttt{weichenxing2023@email.szu.edu.cn}, \texttt{yaoshu@hkust-gz.edu.cn}\\
}

\begin{document}

\maketitle

\begin{abstract}

DeepSeek-R1 has successfully enhanced Large Language Models (LLMs) reasoning capabilities through its rule-based reward system. While it's a ``perfect'' reward system that effectively mitigates reward hacking, such reward functions are often discrete. Our experimental observations suggest that discrete rewards can lead to gradient anomaly, unstable optimization, and slow convergence. To address this issue, we propose \ours{} (\underline{Re}ward \underline{Dit}hering), a method that dithers the discrete reward signal by adding simple random noise. With this perturbed reward, exploratory gradients are continuously provided throughout the learning process, enabling smoother gradient updates and accelerating convergence. The injected noise also introduces stochasticity into flat reward regions, encouraging the model to explore novel policies and escape local optima. Experiments across diverse tasks and different LLMs demonstrate the effectiveness and efficiency of \ours{}. On average, \ours{} achieves performance comparable to vanilla GRPO with only approximately 10\% the training steps, and furthermore, still exhibits a 4\% performance improvement over vanilla GRPO when trained for a similar duration. Visualizations confirm significant mitigation of gradient issues with \ours{}. Moreover, theoretical analyses are provided to further validate these advantages.

\end{abstract}

\section{Introduction}
\label{introduction}

Reinforcement learning (RL) is pivotal in Large Language Models (LLMs) development~\cite{AIAtMeta2024Llama,Anthropic2024Claude,openai2024gpt4technicalreport, wei-etal-2025-flexora}. Initially, RL from human feedback (RLHF)~\cite{DeepReinforcementLearningFromHumanPreferences,ziegler2019finetuning} was employed to align pre-trained LLMs with human preferences~\cite{lang-etal-2024-fine,ouyang2022training}. This typically involves training a separate reward model (RM) on human preference data~\cite{kaufmann2024surveyreinforcementlearninghuman}, which then guides the LLM policy optimization~\cite{ lambert2025reinforcementlearninghumanfeedback}. While effective, this approach introduces considerable training overhead~\cite{Cao_2024}. Subsequently, methods like Direct Preference Optimization (DPO)~\cite{rafailov2023direct, wei2025testtimepolicyadaptationenhanced} were developed, enabling LLMs to learn directly from preference data and thus bypassing explicit RM training. However, these methods still require extensive collection of high-quality preference data. For reasoning tasks such as mathematics and coding, DeepSeek-R1~\cite{deepseekai2025deepseekr1incentivizingreasoningcapability} with Group Relative Policy Optimization~\cite{shao2024deepseekmathpushinglimitsmathematical}(GRPO) proposes an alternative: optimizing the LLM policy directly using a rule-based reward system~\cite{ActiveRewardLearning, wang2025crowdvlmr1expandingr1ability}, thereby avoiding the need for external RMs or large preference datasets. For instance, such a system might assign a reward of 1 for outputs meeting predefined criteria (e.g., correctness, format compliance) and 0 otherwise~\cite{deepseekai2025deepseekr1incentivizingreasoningcapability}. The simplicity and unbiased nature of these rule-based rewards prevent LLMs from hacking them, potentially fostering enhanced reasoning capabilities~\cite{chan2023visionlanguage}.

\begin{figure*}[t]
\centering
\includegraphics[width=1.0\textwidth]{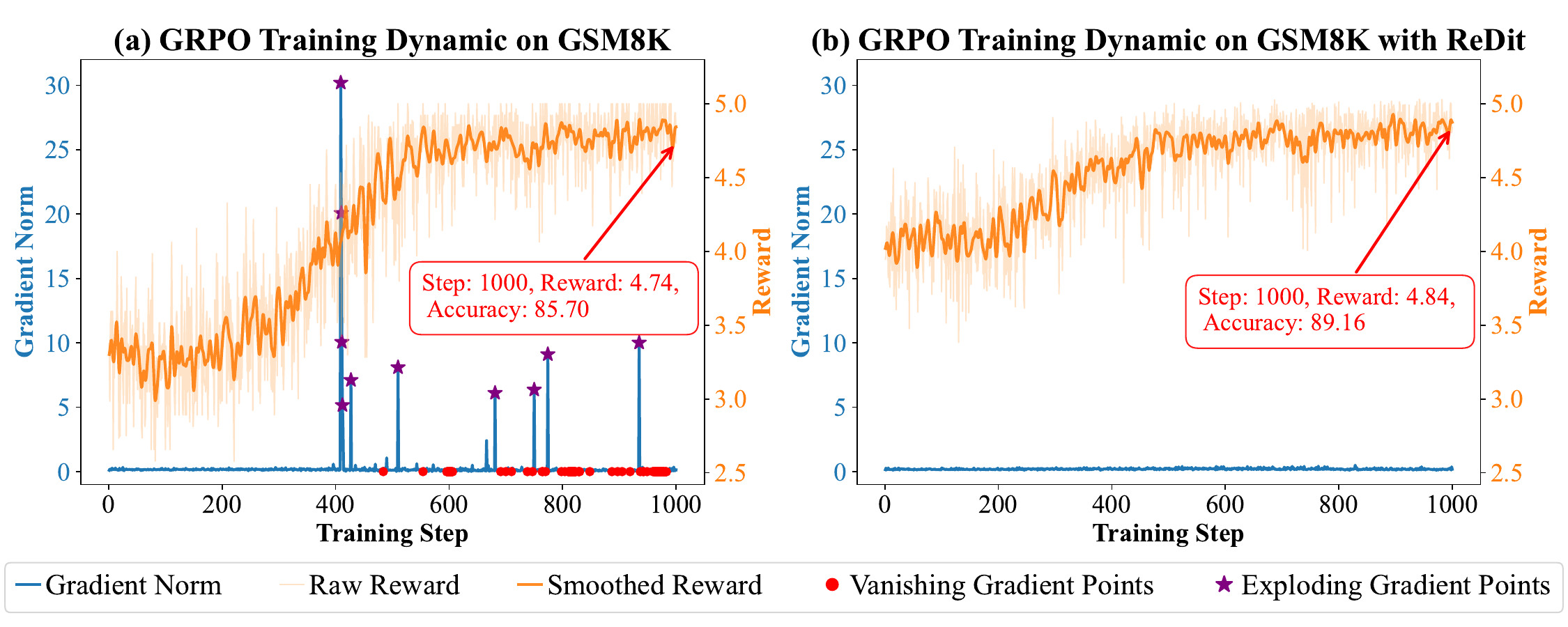}
\caption{Training Dynamics of Gradient Norm and Reward for Qwen2.5-7B~\cite{qwen2025qwen25technicalreport} on GSM8K~\cite{cobbe2021trainingverifierssolvemath} Dataset. The left and right figures compare original gradient norm (before gradient clipping \cite{Zhang2020Why}) and reward trends across training steps. The original GRPO method (the left figure) suffers from significant gradient instability—both vanishing (red dots, norms < 0.01) and exploding (purple asterisks, norms > 5). In contrast, \ours{} with Gaussian reward smoothing (the right figure) effectively stabilizes optimization throughout training.}
\label{fig:Training Dynamics}
\end{figure*}

However, such reward functions are often discrete, posing significant optimization challenges~\cite{rengarajan2022reinforcement, vasan2024revisitingsparserewardsgoalreaching,reward_shaping}. Consider an RL scenario with a binary reward~\cite{once-per-episodeFeedback}: a policy model receives 1 for a correct answer and 0 otherwise. During early training phases, a policy LLM rarely generates completely correct answers, resulting in predominantly zero rewards across mini-batches~\cite{cao-etal-2024-enhancing}. Although the model may engage in exploratory behavior on difficult examples, the corresponding gradients remain minimal due to small advantage magnitudes~\cite{chan2024dense}. Thus, these hard examples and potentially beneficial explorations~\cite{chan2024dense} are largely unexploited during the early stages. Conversely, the model may repeatedly reinforce easy examples~\cite{xie2024textreward}, thus reducing incentives to explore alternative strategies for more difficult problems~\cite{gradient-basedReinforcementLearning}. This phenomenon can lead to training stagnation in intermediate and advanced stages. Consistent with this, as shown in Fig.~\ref{fig:Training Dynamics} (the left figure), we observe that the policy model frequently suffers from gradient vanishing~\cite{razin2024vanishing,GradientMonitoredReinforcementLearning} or explosion~\cite{Evolution-guided, zhang2025gvpogroupvariancepolicy} during these phases. This combination of insufficient exploration and gradient instability substantially impedes model convergence, representing a critical obstacle to efficient RL in LLMs.

\begin{wrapfigure}{r}{0.43\textwidth}  
    \centering
    \includegraphics[width=0.4\textwidth]{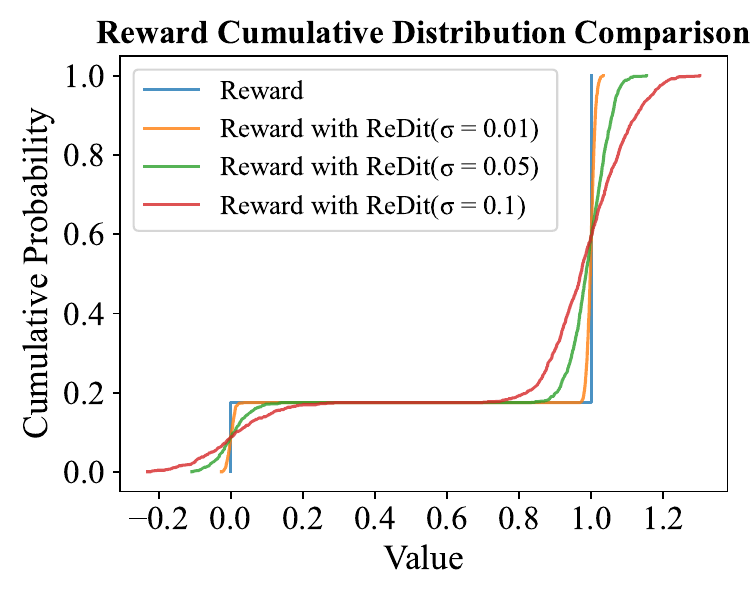}  
    \caption{The figure illustrates how \ours{} of different variances gradually smooth the reward distribution, showing the smoothing effect of perturbations of different variances on the reward distribution.}
    \label{fig:reward_smoothing}
\end{wrapfigure}

This observed phenomenon highlights that even perfectly accurate discrete reward functions face significant limitations within gradient-based optimization frameworks. Lending theoretical support to this, recent studies~\cite{ivison2024unpacking, chen-etal-2024-accuracy, wen2025rethinking} have established that a singular focus on increasing reward model accuracy does not necessarily translate to enhanced language model performance. In particular, \cite{wen2025rethinking} theoretically substantiates the necessity for effective reward models to integrate adequate variance and uncertainty to enable efficient optimization. The theoretical details are given in Sec.~\ref{sec:Theoretical_Principles}. Consequently, we believe that an excellent reward system should achieve a careful equilibrium between correctness and sufficient variance.

Inspired by these observations and theoretical insights, we propose \ours{}, a simple but effective technique that introduces zero-mean random perturbations to discrete reward signals during training. By adding small random noise to the reward function (Fig.~\ref{fig:reward_smoothing}), \ours{} effectively softens the original hard reward function. Compared to a hard reward function, a softened one can provide greater reward variance within mini-batches, which, as indicated by previous research, can lead to enhanced model performance and accelerated convergence.

Fig.~\ref{fig:Training Dynamics} (the right figure) illustrates the impact of our proposed \ours{} on LLM policy optimization for the GSM8K~\cite{cobbe2021trainingverifierssolvemath} task.  The orange lines indicate that during early training phases, GRPO with \ours{} achieves significantly higher average rewards compared to the baseline (GRPO without \ours{}), demonstrating the efficacy of our method. We hypothesize that \ours{} encourages broader exploration by assigning varied rewards to outputs that only partially meet strict evaluation criteria, thereby accelerating convergence. Towards the end of training (1000 steps), while both policy models attain high rewards, our approach with \ours{} exhibits superior performance on the test set, indicating enhanced generalization. Additionally, \ours{} demonstrates more robust gradient updating. As shown in Fig.~\ref{fig:Training Dynamics} (the left figure), phenomena such as gradient vanishing (red point) and explosion (purple star) emerge during training with the baseline. In contrast, GRPO with \ours{} maintains stable gradients throughout the training process. 
These findings highlight the advantages of \ours{}: more stable policy optimization, faster convergence, and improved overall performance.

Moreover, theoretical analysis indicates that a greater reward variance can enhance performance and accelerate convergence in reinforcement learning \cite{wen2025rethinking}. We increase reward variance within mini-batches while preserving the expected gradient through reward dithering. By carefully injecting noise into the reward function, \ours{} achieves a balance between reward signal fidelity and reward variance, leading to enhanced policy optimization.

In summary, our main contributions are:
\begin{itemize}
    \item We observe that policy optimization under discrete reward functions suffer from unstable gradients and slow convergence(Section~\ref{sec:discrete_reward_challenges}).
    \item We propose Reward Dithering (\ours{}), a simple yet effective technique that introduces perturbations to discrete rewards. This method is shown to accelerate convergence speed and enhance final model performance (Algorithm~\ref{alg:reward_smoothing} and Section~\ref{sec:ours_method}).
    \item Extensive experiments across diverse downstream tasks, reinforcement learning algorithms, and perturbation distributions demonstrate that \ours{} achieves superior performance and enhanced convergence (Section~\ref{sec:results}).
    \item Theoretical analysis proves that \ours{} produces an unbiased estimate of the original gradient (Proposition~\ref{prop:unbiased}) and introduces beneficial gradient variance that mitigates vanishing and exploding gradients (Proposition~\ref{prop:variance}).
\end{itemize}

\section{Preliminaries}
\label{sec:grpo_background}

We frame LLM generation as a sequential decision-making problem solvable via RL. The process is modeled as a Markov Decision Process (MDP)~\cite{hallak2015contextualmarkovdecisionprocesses} where the state $s_t = q; o_{<t}$ includes the prompt $q$ and generated tokens $o_{<t}$, the action $o_t$ is the next token selected from the vocabulary, and the policy $\pi_\theta(o_t | s_t)$ is parameterized by $\theta$. The goal is to optimize the policy to maximize the expected sequence-level reward $R(q, o) = \sum_{t=1}^{|o|} r(s_t, o_t)$ over the prompt distribution $p_Q$:
\begin{equation}
    J(\pi_\theta) = \mathbb{E}_{q \sim p_Q} \left[ \mathbb{E}_{o \sim \pi_\theta(\cdot|q)}[R(q, o)] \right].
    \label{eq:rl_objective}
\end{equation}

Recently, GRPO~\cite{shao2024deepseekmathpushinglimitsmathematical} was proposed as a PPO alternative that eliminates the need for independent RMs and value functions. GRPO typically processes sparse, discrete rewards directly, rather than continuous RM scores. For tasks like mathematical reasoning, this discrete reward $R(q, o) \in \{0, 1\}$ is often determined by a simple function checking correctness or format. GRPO estimates the advantage $\hat{A}_{i,t}^{\text{GRPO}}$ by sampling $G$ responses $\{o_i\}_{i=1}^G$ and normalizing their discrete rewards within the set. Its objective function, which includes a KL divergence term $D_{\text{KL}}(\pi_\theta || \pi_{\text{ref}})$ for stability, is given by:
\begin{equation}
\begin{aligned}
J_{\text{GRPO}}(\theta) = \mathbb{E}_{q \sim p_Q} \left[ \frac{1}{G} \sum_{i=1}^G \sum_{t=1}^{|o_i|} \min \left( r_{i,t}(\theta) \hat{A}_{i,t}^{\text{GRPO}}, \text{clip} \left( r_{i,t}(\theta), 1 - \epsilon, 1 + \epsilon \right) \hat{A}_{i,t}^{\text{GRPO}} \right) \right] \\
- \beta \mathbb{E}_{q \sim p_Q} \left[ D_{\text{KL}}(\pi_\theta(\cdot|q) || \pi_{\text{ref}}(\cdot|q)) \right],
\label{eq:grpo_objective}
\end{aligned}
\end{equation}
where $r_{i,t}(\theta) = \frac{\pi_\theta(o_{i,t}|s_{i,t})}{\pi_{\theta_{\text{old}}}(o_{i,t}|s_{i,t})}$. Subsequent methods such as DAPO~\cite{yu2025dapoopensourcellmreinforcement}, Dr.GRPO~\cite{liu2025understandingr1zeroliketrainingcritical}, and REINFORCE++~\cite{hu2025reinforceefficientrlhfalgorithm} generally adopt this discrete reward paradigm (see Appendix~\ref{sec:related_work} for more related work). While simplifying the overall RL process by avoiding complex RMs, this shift to discrete, sequence-level rewards introduces significant optimization challenges. The inherent sparsity and abrupt value changes (e.g., 0 to 1) hinder policy gradient estimation and lead to training instability (Section~\ref{sec:discrete_reward_challenges}). 

\section{Motivation}
This section articulates the fundamental motivations driving our research and establishes the critical challenges that our work aims to address. In Section~\ref{sec:discrete_reward_challenges}, we examine the optimization challenges inherent in discrete reward structures, followed by an exposition of the theoretical principles informing our methodological framework in Section~\ref{sec:Theoretical_Principles}.
\subsection{Difficulties in Optimization Caused by Discrete Rewards}
\label{sec:discrete_reward_challenges}
\begin{figure*}[t]
\centering
\includegraphics[width=1.0\textwidth]{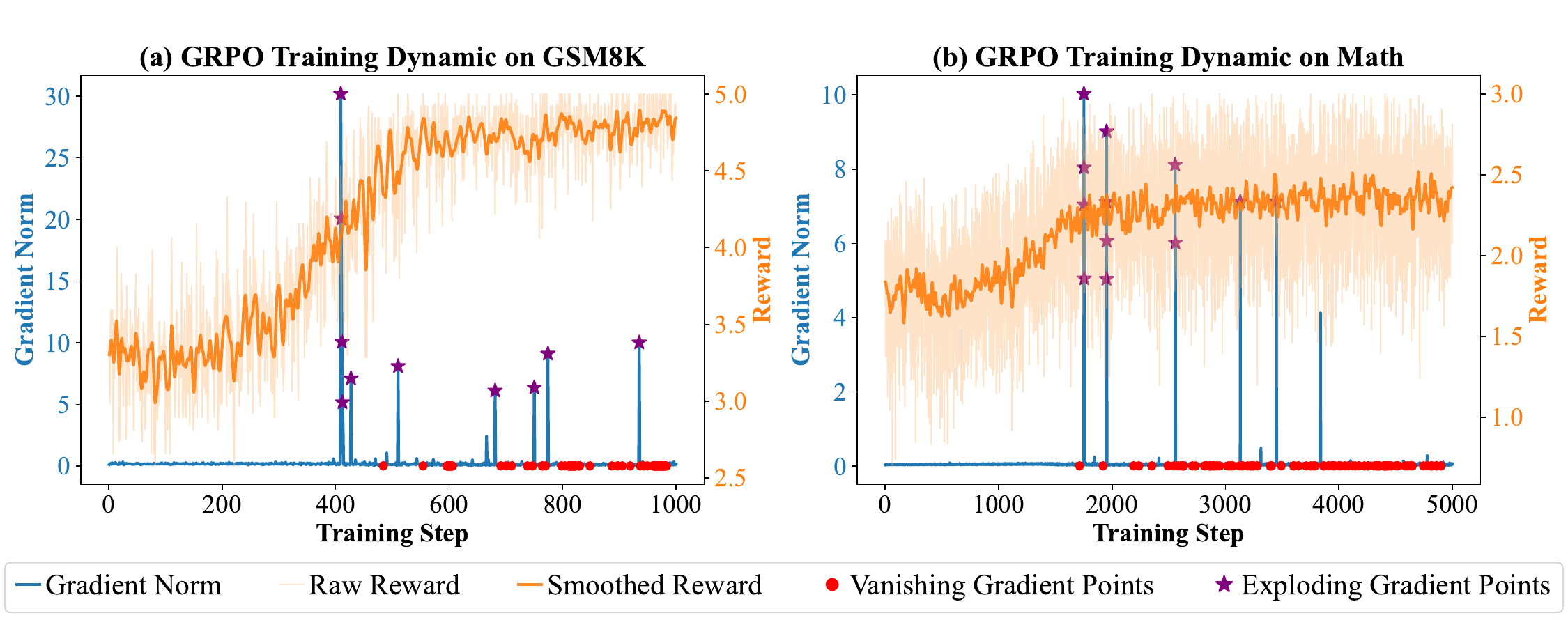}
\vspace{-5mm}
\caption{Qwen2.5-7B~\cite{qwen2025qwen25technicalreport} Gradient norm and reward training dynamics of standard GRPO on GSM8k and MATH datasets. During the whole optimization process, the gradient of standard GRPO is unstable, and there are a lot of gradient vanishing or gradient exploding cases.}
\label{fig:Training Dynamics GRPO}
\vspace{-3mm}
\end{figure*}

Optimizing LLM policies using algorithms like GRPO in conjunction with discrete sequence-level rewards (e.g., binary correctness metrics) presents significant optimization challenges. Fig.~\ref{fig:Training Dynamics GRPO} plots the policy gradient norm (blue line) and average reward (orange line) during standard GRPO training on the GSM8K and MATH datasets, respectively. Two main issues are immediately apparent:
\begin{wrapfigure}{r}{0.4\textwidth}  
    \centering
    \includegraphics[width=0.4\textwidth]{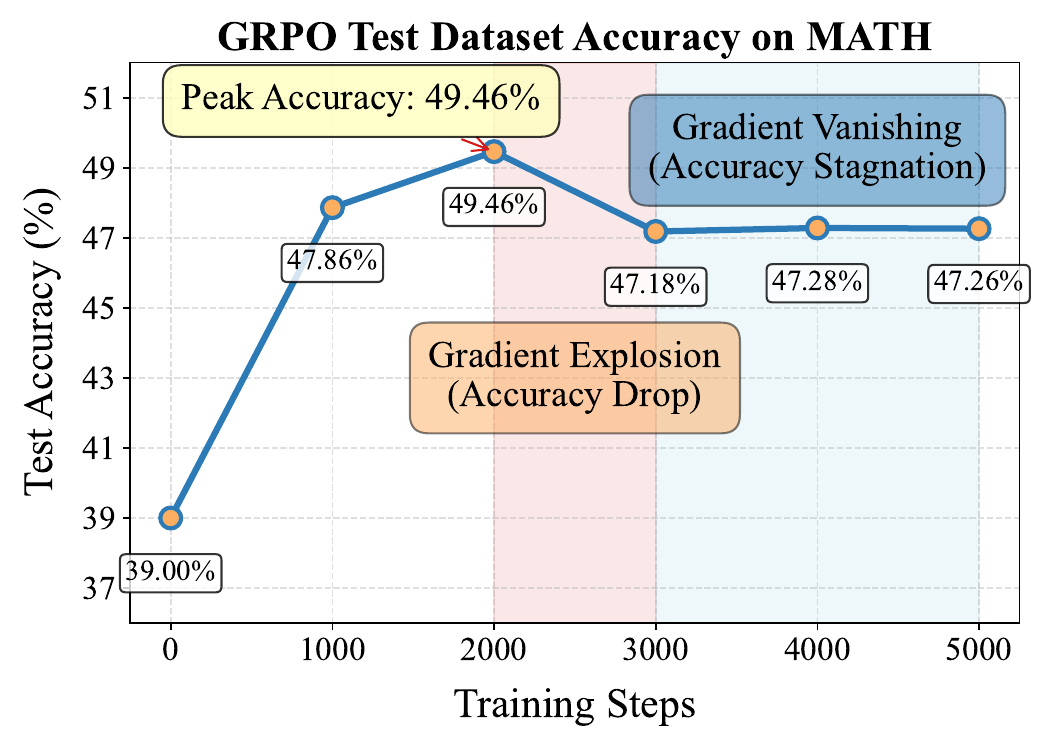}  
    \vspace{-3mm}
    \caption{GRPO has unstable performance on the MATH test set. The figure plots the test accuracy achieved for the checkpoints saved during the training run shown in Fig.~\ref{fig:Training Dynamics GRPO}(the right figure).}
    \label{fig:math_test_accuracy}
    \vspace{-3mm}
\end{wrapfigure}
\textbf{Gradient Vanishing.}  The figure illustrates instances where the gradient norm approaches zero (red dot), occurring when most examples in a GRPO batch yield identical binary rewards. Consequently, the population relative advantage estimate $\hat{A}_{i,t}^{\text{GRPO}}$ becomes negligible across examples, providing insufficient learning signals and causing training stagnation. This phenomenon is evident in Fig.~\ref{fig:Training Dynamics GRPO}(the right figure) post-step 2000.

\textbf{Gradient Explosion.} Conversely, training dynamics exhibit sporadic sharp spikes in gradient norm (purple asterisks) when small policy changes cause sequences to transition from incorrect (reward 0) to correct (reward 1). These transitions create disproportionately large advantage estimates for newly successful sequences, triggering sudden, destabilizing gradient updates as shown in Fig.~\ref{fig:Training Dynamics GRPO}(the left figure). Such spikes induce reward fluctuations in subsequent steps, hindering smooth convergence and learning efficiency.

The discrete, sparse rewards induce unstable oscillations between vanishing and exploding gradients. Fig.~\ref{fig:math_test_accuracy} demonstrates that model performance fluctuates correspondingly with these oscillations. This inherent instability not only compromises optimization efficiency but also serves as a key motivation for our research.

\subsection{Theoretical Principles to Address the Limitations of Discrete Rewards}
\label{sec:Theoretical_Principles}
To overcome the critical challenges with discrete rewards outlined in Section~\ref{sec:discrete_reward_challenges}, we propose a direct approach to enhance reward signal quality. Our solution derives from key theoretical insights in policy optimization, particularly Theorem~\ref{thm:lower_bound} and Theorem~\ref{thm:upper_bound}, which reveal fundamental relationships between reward variance, model accuracy, and learning efficiency. See Appendix~\ref{definition} for definitions.

\begin{tcolorbox}[colback=red!5!white,colframe=red!70!white, left=0.5mm, right=1mm, top=1mm, bottom=1mm]
\begin{theorem}[Policy network optimization time lower bound]
\label{thm:lower_bound}
\textit{\fontfamily{ppl}\selectfont
From Theorem 1 in \cite{razin2025makesrewardmodelgood}. Suppose that we maximize the objective (Eq.~\eqref{eq:rl_objective}), using a general autoregressive policy $\pi_\theta (\ybf | \xbf) = \prod_{l = 1}^{\ybf} \smax ({ \nn_\theta (\xbf, \ybf_{< l}) }_{ \ybf_l })$. 
For any $\gamma > 0$, prompt $\xbf \in \X$, and reward function $r$, the time it takes until $\EE\nolimits_{\ybf \sim \pi_{\theta (t)} (\cdot | \xbf)} [r(\xbf,\ybf)] \geq \EE\nolimits_{\ybf \sim \pi_{\theta (0)} (\cdot | \xbf)} [r (\xbf, \ybf)] + \gamma$ is:
\begin{equation}
\Omega \left( \EE\nolimits_{\xbf' \sim \datasetpg} \left[ \var_{\ybf \sim \pi_{\theta(0)} (\cdot | \xbf')} (r (\xbf', \ybf)) \right]^{-\frac{1}{3}} \right)
\end{equation}
The reward variance is: $\var_{\ybf \sim \pi_\theta (\cdot | \xbf)} [ r (\xbf, \ybf) ] := \EE\nolimits_{\ybf \sim \pi_{\theta} (\cdot | \xbf) } [ [ r (\xbf, \ybf) - \EE\nolimits_{\ybf' \sim \pi_{\theta} (\cdot | \xbf)} [r (\xbf, \ybf') ] ]^2 ]$.}
\end{theorem}
\end{tcolorbox}
Theorem~\ref{thm:lower_bound} establishes that the time $t_\gamma$required for policy improvement is inversely proportional to reward variance. When rewards exhibit insufficient variance—failing to adequately differentiate between high-quality and low-quality outputs under policy $\pi_\theta$, convergence slows significantly. This finding suggests that strategically increasing reward variance can accelerate policy convergence.
\begin{tcolorbox}[colback=red!5!white,colframe=red!70!white, left=0.5mm, right=1mm, top=1mm, bottom=1mm]
\begin{theorem}[Policy network optimization time upper bound]
\label{thm:upper_bound}
\textit{\fontfamily{ppl}\selectfont
From Theorem 2 in \cite{razin2025makesrewardmodelgood}.
Assume $\pi_{\theta}$ is a policy of the form $\pi_\theta (\ybf | \xbf) = \smax [ \theta_{: , \xbf} ]_{\ybf}$. Given a prompt $\xbf \in \datasetpg$, let $\gamma > 0$ and denote by $t_\gamma > 0$ the initial time of $\EE\nolimits_{\ybf \sim \pi_{\theta (t)} (\cdot | \xbf)} [ \gtreward (\xbf, \ybf) ] \geq \EE\nolimits_{\ybf \sim \pi_{\theta (0)} (\cdot | \xbf)} [ \gtreward (\xbf, \ybf) ] + \gamma$.
For any initial policy $\pi_{\theta (0)}$, a perfect RM converges to $t_\gamma$ that can be arbitrarily large, while a relatively inaccurate RM has an upper bound of $\OO (\pi_{\theta (0)} (y^\gamma | \xbf)^{-1} )$.}
\end{theorem}
\end{tcolorbox}
Complementarily, Theorem~\ref{thm:upper_bound} demonstrates that effective reward models must incorporate a calibrated degree of uncertainty. This controlled uncertainty creates essential exploration space during early training stages, preventing premature convergence and facilitating more efficient optimization.

While perfectly accurate reward functions resist reward hacking, they paradoxically impede optimization by producing discrete rewards with minimal variance and insufficient randomness. This limitation severely constrains the growth rates of both training reward $r_{RM}$ and true reward $r_G$ during policy gradient updates. To address this fundamental tension, we introduce \ours{}—a method that injects zero-mean perturbations into discrete rewards. This approach preserves the expected reward value while introducing beneficial variance and controlled uncertainty in each update step, dramatically improving both model performance and convergence speed.

\section{Reward Dithering (\ours{}) }
\label{sec:ours_method}
As discussed previously, the discrete nature of rewards commonly used in GRPO can lead to unstable gradient dynamics. To address this, we propose \textbf{\ours{} }. The core idea, detailed in Algorithm~\ref{alg:reward_smoothing}, is to inject calibrated, zero-mean perturbations into the discrete rewards obtained from sampled outputs before using them to compute the GRPO objective for policy updates. Importantly, our \ours{} method preserves the overall optimization structure of the GRPO objective function as defined in Eq.~\eqref{eq:grpo_objective}, the optimization still aims to maximize this objective.

The crucial modification introduced by \ours{} lies in \textit{how} the advantage term $\hat{A}_{i,t}^{\text{GRPO}}$ within Eq.~\eqref{eq:grpo_objective} is computed. Instead of directly using the raw discrete rewards $r_i = r(o_i)$ obtained for each sampled output $o_i$ in the batch $\{o_i\}_{i=1}^G$ (line 3 in Algorithm~\ref{alg:reward_smoothing}), we first compute \textbf{smoothed rewards} $\tilde{r}_i$. This is done by adding independently sampled zero-mean perturbation $\epsilon_i$ (e.g., from $\mathcal{N}(0, \sigma^2)$ or $\mathcal{U}[-a, a]$) to each discrete reward (line 4 in Algorithm~\ref{alg:reward_smoothing}):
\begin{equation}
    \tilde{r}_i = r_i + \epsilon_i
    \label{eq:smoothed_reward}
\end{equation}
These smoothed rewards $\{\tilde{r}_k\}_{k=1}^G$ are then used as the basis for calculating the advantage. GRPO often computes advantage based on the relative performance within the batch, typically involving normalization. With \ours{}, the core component of the advantage calculation, which relies on these rewards, is effectively modified as follows:
\begin{equation}
    \hat{A}_{i,t}^{\text{GRPO}} \propto \frac{r_i - \text{mean}(\{{r}_k\}_{k=1}^G)}{\text{std}(\{{r}_k\}_{k=1}^G)} \quad \xrightarrow{\text{\ours{}}} \quad \hat{A}_{i,t}^{\text{Dithering}} \propto \frac{\tilde{r}_i - \text{mean}(\{\tilde{r}_k\}_{k=1}^G)}{\text{std}(\{\tilde{r}_k\}_{k=1}^G)}
    \label{eq:advantage_modification}
\end{equation}
Thus, the relative standing of each output $o_i$ within the batch, which informs its advantage $\hat{A}_{i,t}^{\text{GRPO}}$ used in Eq.~\eqref{eq:grpo_objective}, is determined by the continuous smoothed reward $\tilde{r}_i$ rather than the discrete $r_i$. This substitution transforms the optimization landscape. By introducing continuous variations via $\tilde{r}_i$, the added noise provides informative, non-zero gradients even when discrete rewards $r_i$ are sparse or identical within a batch, mitigating gradient vanishing. It also dampens the sharp changes in expected advantage resulting from small policy shifts affecting discrete outcomes, thus reducing the likelihood of gradient explosion. This overall smoothing effect facilitates a more stable gradient flow, enabling more robust and efficient optimization of the policy $\pi_\theta$ using the GRPO objective (line 5 in Algorithm~\ref{alg:reward_smoothing}).
\begin{algorithm}
\caption{\ours{} within one optimization step}
\label{alg:reward_smoothing} 
\begin{algorithmic}[1] 
    \STATE {\bfseries Input:} Base policy $\pi_{\theta_{\text{old}}}$; Discrete reward function $r: \mathcal{O} \to \{0, 1,2,3,...\}$; Prompt $q$; Number of samples $G$. Noise parameters: Gaussian std dev $\sigma > 0$ \textbf{or} Uniform radius $a > 0$.
    \STATE {\bfseries Output:} Updated policy $\pi_{\theta}$.
    \vspace{0.5em} 
    \STATE Sample $G$ outputs $\{o_i\}_{i=1}^G \sim \pi_{\theta_{\text{old}}}(\cdot \mid q)$ and compute $r_i \leftarrow r(o_i)$ for $i=1, \dots, G$.
    \STATE Sample $\epsilon_i \sim \mathcal{N}(0, \sigma^2)$ \textbf{or} $\mathcal{U}[-a, a]$ and compute $\tilde{r}_i \leftarrow r_i + \epsilon_i$ for $i=1, \dots, G$.\COMMENT{Generate noise and smooth rewards.}
    \STATE Compute $J_{\text{GRPO}}$ using $\{\tilde{r}_i\}_{i=1}^G$ and $\theta \leftarrow \text{Optimize}(\theta_{\text{old}}, J_{\text{GRPO}},\tilde{r}_i)$.\COMMENT{Optimization}
    \STATE {\bf return} Updated policy $\pi_{\theta}$.
\end{algorithmic}
\end{algorithm}

\section{Empirical Results}
\label{sec:results}

This section presents a thorough evaluation of our \ours{} framework, assessing its effectiveness and efficiency. We begin by detailing the datasets and experimental configurations in Section~\ref{sec:experimental settings}. Subsequently, Section~\ref{sec:main results} provides a comprehensive analysis of the primary findings. To isolate the contributions of key components, we also conduct ablation studies, the results of which are presented in Section~\ref{sec:ablation study}.

\begin{wraptable}{r}{0.6\textwidth} 
\small
    \centering
    \caption{Comparison of the mean and variance of accuracy for different baselines under 9000 steps on GSM8K.} 
    \label{tab:result-2}
    \resizebox{0.6\textwidth}{!}{
        \begin{tabular}{@{}lccc@{}} 
            \toprule
            \textbf{Name} & \textbf{DAPO} & \textbf{DR.GRPO} & \textbf{REINFORCE++} \\
            \midrule
            Baseline & 87.52 & 86.13 & 86.25 \\
            w/ \textbf{ours(Gauss)} & \textbf{89.34} ($\pm$ 0.04)& \textbf{87.69} ($\pm$ 0.06)& \textbf{87.96} ($\pm$ 0.03)\\
            w/ \textbf{ours(Uniform)} & 88.57 ($\pm$ 0.01) & 87.34 ($\pm$ 0.07) & 87.59 ($\pm$ 0.09) \\
            \cmidrule(lr){1-4} 
            \multirow{1}{*}{$\Delta$} & +1.82 & +1.56 & +1.71 \\
            \bottomrule
        \end{tabular}
    }
\end{wraptable}

\subsection{Datasets and Setup}
\label{sec:experimental settings}
To rigorously evaluate the effectiveness of our proposed \ours{} framework, we conducted extensive experiments. The specific experimental settings are detailed below.

\textbf{Datasets.} Our dataset selection and setup largely follow the methodology of~\cite{shao2024deepseekmathpushinglimitsmathematical}, primarily to assess the mathematical reasoning capabilities of the models. This encompasses mathematical problem-solving datasets such as GSM8K~\cite{cobbe2021trainingverifierssolvemath} and MATH~\cite{hendrycks2021measuringmathematicalproblemsolving}, as well as the multimodal geometric reasoning dataset Geometry3K~\cite{lu2021intergpsinterpretablegeometryproblem}. Each dataset provides distinct training and test splits, which we utilize accordingly for model training and subsequent evaluation. See the Appendix~\ref{sec:dataset} for details of the dataset.

\textbf{Reward Functions.} We designed dataset-specific reward functions. For the GSM8K dataset, which involves simpler problem structures, we implemented several reward types: accuracy-based, strict format adherence, sort format adherence, integer value correctness, and inference step adherence. For the more complex MATH and Geometry3K datasets, our supervision relied solely on accuracy-based and inference-based reward functions. Detailed implementations of these reward functions are provided in the Appendix~\ref{sec:reward}.

\textbf{Initial Policy}. To rigorously assess the effectiveness of \ours{} without confounding factors introduced by supervised fine-tuning (SFT), we initialized our experiments directly with instruct models without any additional SFT training. Previous research by~\cite{shao2025spuriousrewardsrethinkingtraining} demonstrated that even random rewards can enhance performance for Qwen models. Therefore, we conducted comprehensive evaluations across a diverse set of instruction-tuned models, including Qwen2.5-7B-Instruct, Qwen2.5-VL-7B-Instruct \cite{qwen2025qwen25technicalreport}, Llama-3.2-3B-Instruct, Llama-3.1-8B-Instruct \cite{AIAtMeta2024Llama}, Ministral-8B-Instruct-2410, and Mistral-7B-Instruct-v0.3 \cite{jiang2023mistral7b}, to establish the generalizability of \ours{}.

\textbf{Random Seeds}. Our method incorporates random sampling, which can introduce variance to the experimental outcomes. To thoroughly assess the impact of this stochasticity and ensure the robustness of our results, we executed all main experiments using multiple distinct random seeds. Specifically, we selected five seeds: None (no seed), 42, 123, 888, 2025, and 9999. All metrics reported in our final results represent the mean and variance computed across these five independent runs.

\textbf{Other Training Settings.} For parameter-efficient fine-tuning, we employed Low-Rank Adaptation (LoRA)~\cite{hu2022lora}. Our implementation leverages the official GRPO implementation within the TRL library~\cite{vonwerra2022trl}. Specific configurations for LoRA and GRPO parameters are detailed in the Appendix~\ref{sec:parameters}. Model evaluation was conducted using the OpenCompass~\cite{2023opencompass}. All experiments were executed on one NVIDIA H20 GPU.

\subsection{Main Results}
\label{sec:main results}

\begin{wraptable}{r}{0.6\textwidth} 
\small
    \centering
    \caption{Test mean and variance of accuracy comparison across datasets for original Backbone, GRPO, and \ours{}.} 
    \label{tab:result-1}
    \resizebox{0.6\textwidth}{!}{
        \begin{tabular}{@{}lccc@{}} 
            \toprule
            \textbf{Name} & \textbf{GSM8K} & \textbf{MATH} & \textbf{Geometry3K} \\
            \midrule
            Backbone & 84.91 & 39 & 40.43 \\
            GRPO(Baseline) & 89.07 & 48.01 & 43.10 \\
            w/ \textbf{ours(Gauss)} & \textbf{90.76} ($\pm$ 0.06)& \textbf{52.55} ($\pm$ 0.03)& \textbf{44.67} ($\pm$ 0.03)\\
            w/ \textbf{ours(Uniform)} & 90.46 ($\pm$ 0.07)& 51.96 ($\pm$ 0.06)& 44.36 ($\pm$ 0.04)\\
            \cmidrule(lr){1-4} 
            $\Delta$ & +1.69 & +4.54 & +1.57 \\
            \bottomrule
        \end{tabular}
    }
\end{wraptable}
In our main experiments, we validate the effectiveness of our proposed \ours{}. For these experiments, we primarily use either a uniform smoothing kernel with radius $a = 0.05$  or a Gaussian smoothing kernel with standard deviation $\sigma = a/\sqrt{3}$. More experimental results can be found in the Appendix~\ref{more_result}.
\begin{figure*}[t]
\centering
\includegraphics[width=1.0\textwidth]{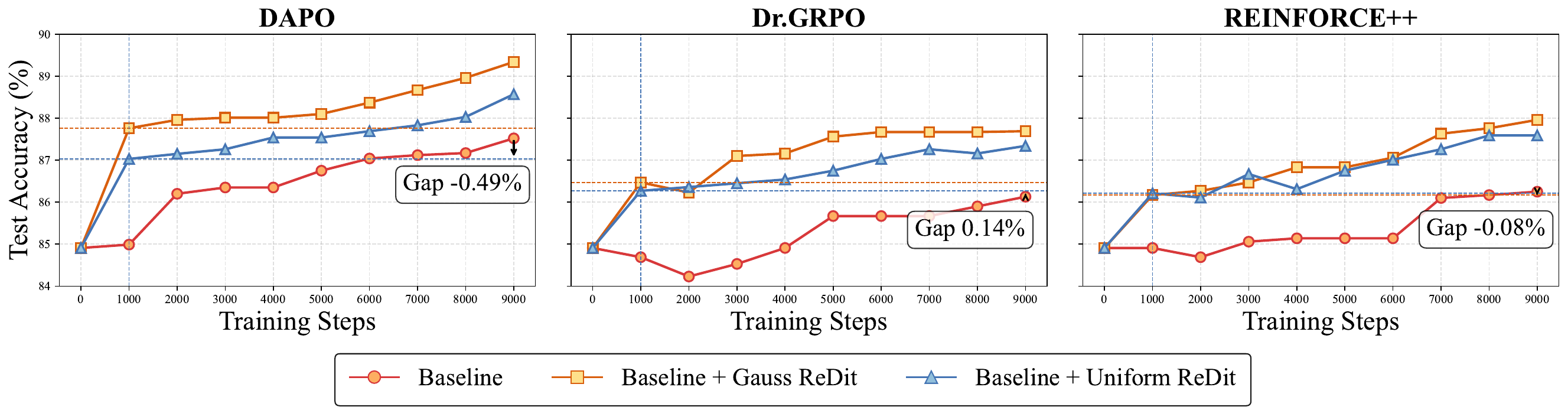}
\caption{Accuracy of different GRPO variants (DAPO, DR.GRPO, REINFORCE++) tested on the GSM8K dataset. The horizontal dashed line highlights the performance of using \ours{} at about 1000 training steps, and even after 9000 steps, its accuracy is comparable to the baseline.}
\label{fig:result-2}
\end{figure*}

\begin{figure*}[t]
\centering
\includegraphics[width=1.0\textwidth]{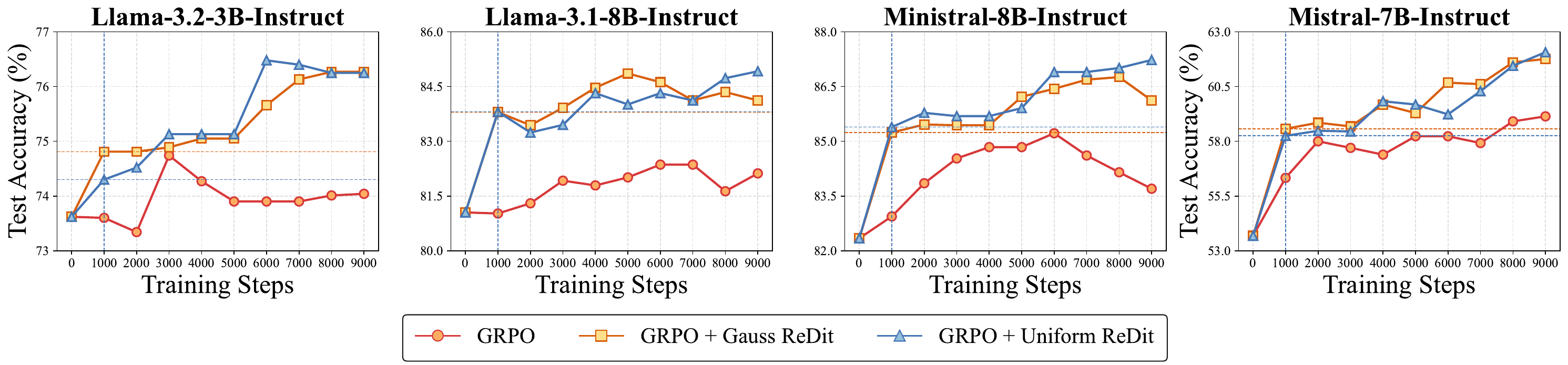}
\caption{Accuracy of different LLMs on GSM8K. \ours{} improves training efficiency and final performance in various LLMs.}
\label{fig:result-3}
\end{figure*}

\begin{figure*}[t]
\centering
\includegraphics[width=1.0\textwidth]{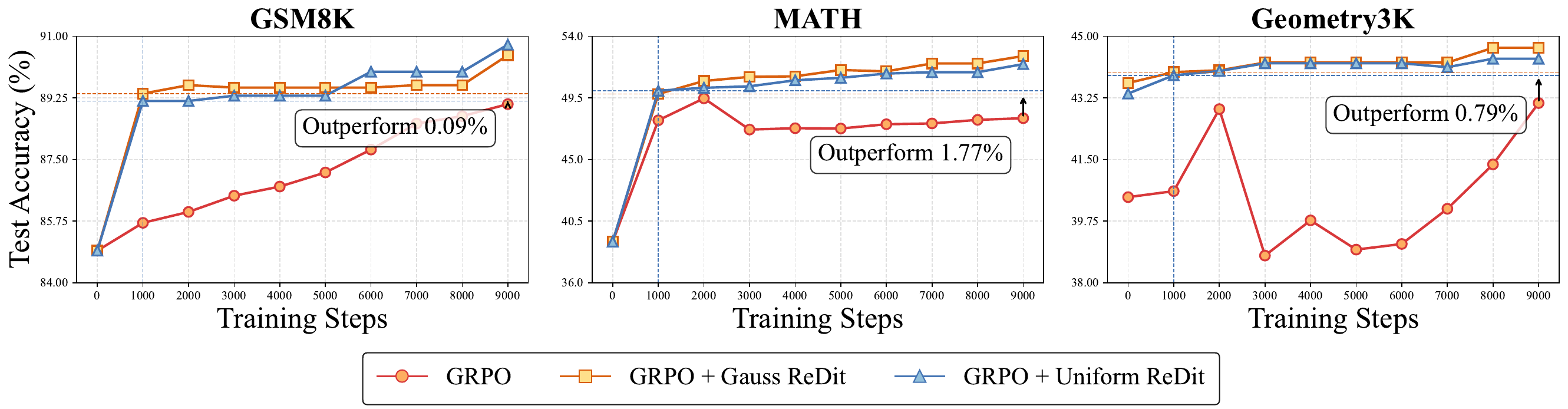}
\caption{Test accuracy across datasets. The horizontal dashed line marks \ours{}'s performance at 1000 steps, which GRPO fails to match even after 9000 steps.}
\label{fig:result-1}
\end{figure*}

\textbf{Accelerated Convergence Across Datasets and LLMs.} 
We demonstrate that integrating our proposed method, ReDit, with GRPO substantially accelerates convergence and improves final performance across a wide range of datasets (Fig.~\ref{fig:result-1}) and LLMs, including Llama-3.2-3B, Llama-3.1-8B, Ministral-8B and Mistral-7B (Fig.~\ref{fig:result-3}). On all tested models, both Gaussian and uniform variants of ReDit enable GRPO to reach a competitive performance level within merely 1000 training steps. Notably, this performance already surpasses that of the baseline GRPO trained for the full 9000 steps. Consequently, ReDit not only enhances training efficiency but also leads to superior final accuracy. The Gaussian variant, in particular, consistently yields the strongest results and promotes more stable training trajectories with lower volatility compared to the baseline. We also present more experimental results in the appendix. The results on the code dataset are detailed in Appendix~\ref{code}, the results on the full parameter fine-tuning method are detailed in Appendix~\ref{full}, and the results on the Deepseek-R1 distillation model are detailed in Appendix~\ref{distillation}.

\textbf{Generalization to Diverse Baselines.} Fig.~\ref{fig:result-2} presents results from applying \ours{} to additional reinforcement learning baselines (DAPO, Dr.GRPO, and REINFORCE++) on the GSM8K dataset. Across all algorithms, \ours{} (both Gaussian and uniform variants) consistently enhances performance and accelerates learning. Beyond these early-stage improvements, \ours{} also substantially boosts the final accuracy of these baselines, as quantitatively demonstrated in Table~\ref{tab:result-2}. These accuracy gains (Table~\ref{tab:result-1}) complement the qualitative evidence in Fig.~\ref{fig:result-2}, confirming that \ours{} enables faster and more stable learning across diverse algorithms.

\begin{figure}[htbp] 
    \begin{minipage}{0.46\textwidth}
        \centering
        \includegraphics[width=1\textwidth]{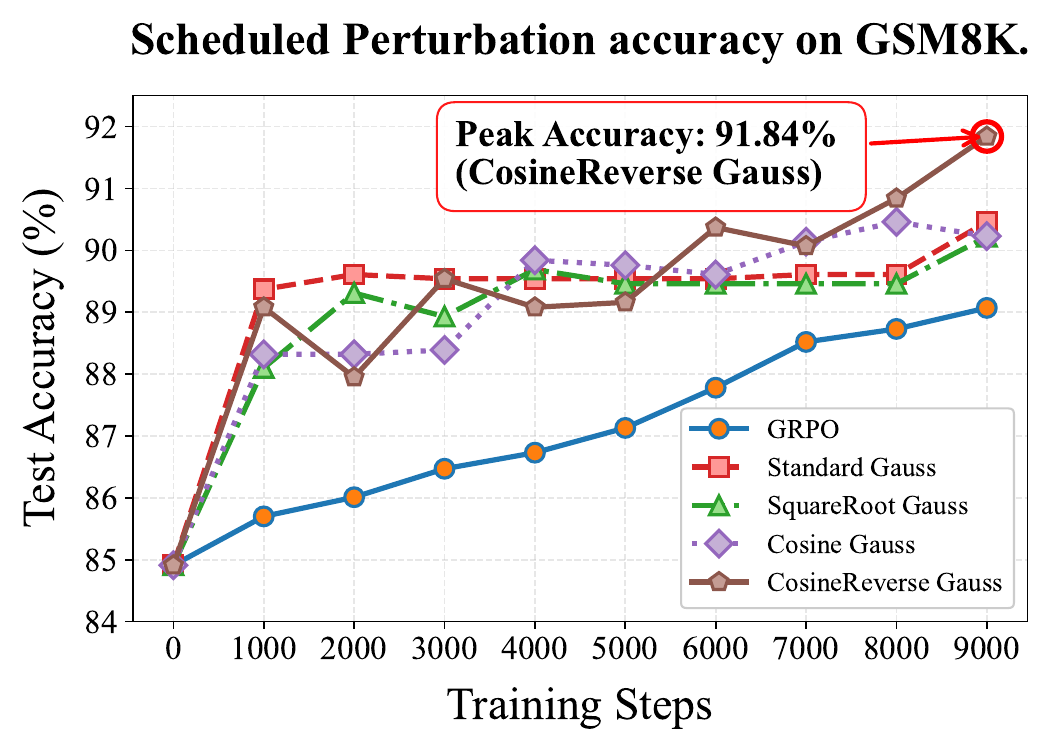}
        \caption{CosineReverse perturbation achieves the best performance.}
        \label{fig:ablation_a_value}
    \end{minipage}
    \hfill
    \begin{minipage}{0.46\textwidth}
        \centering
        \includegraphics[width=1\textwidth]{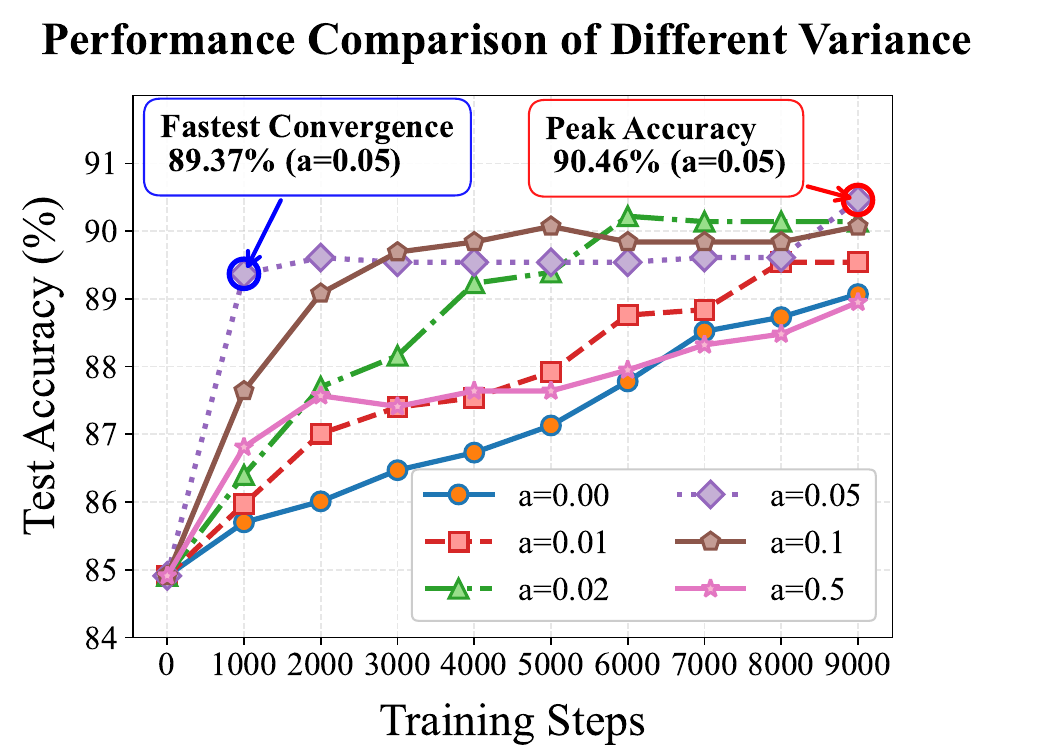}
        \caption{Appropriate perturbation achieves the best performance.}
        \label{fig:schedule-results}
    \end{minipage}
\end{figure}

\textbf{Optimal Performance with Scheduled Perturbation.} We further investigate convergence behavior under various scheduled perturbation schemes: SquareRoot, Cosine, and CosineReverse perturbations. These schedules dynamically adjust perturbation variance throughout training, potentially benefiting model learning. Fig. in the Appendix~\ref{sec: scheduled} illustrates the different perturbation schedules, while Fig.~\ref{fig:ablation_a_value}presents their performance. Compared to standard GRPO, \ours{} achieves both faster convergence and superior final performance, with the CosineReverse perturbation schedule yielding particularly strong results. Additional details are provided in the Appendix~\ref{sec: scheduled}.

\subsection{Ablation Studies}
\label{sec:ablation study}

\textbf{Perturbation variance affects performance. } To study the sensitivity of \ours{} to the perturbation amplitude, we performed an ablation study by varying the parameter $a$ in the Gaussian smoothing kernel with standard deviation $\sigma = a/\sqrt{3}$. This effectively changes the variance of the applied perturbation. As shown in Fig.~\ref{fig:schedule-results}, applying reward smoothing (i.e., for any $a > 0.00$) consistently leads to faster convergence compared to the baseline without smoothing ($a=0.00$). Moreover, in most cases, increasing the perturbation amplitude (larger $a$) tends to improve the final performance of the model. Notably, the configuration with $a=0.05$ shows superior performance, achieving not only the fastest convergence but also the best peak model performance, see the Fig.~\ref{fig:schedule-results} annotation. However, these results highlight a key trade-off. While moderate perturbations are beneficial, excessive perturbations (e.g., $a=0.5$) may over-smooth the reward landscape. This may mask the original reward signal and lead to performance degradation. Conversely, if the perturbation variance is too small (e.g., $a=0.01$), the smoothing effect is small and the improvement over the baseline is limited. This suggests that there is an optimal perturbation variance. We recommend conducting preliminary experiments on a smaller dataset to effectively determine this optimal variance before applying it to larger-scale training scenarios. For a detailed theoretical introduction to $\sigma$, please refer to Section~\ref{sec:theoretical}.

\begin{figure}[htbp] 
    \begin{minipage}{0.46\textwidth}
        \centering
        \includegraphics[width=1\textwidth]{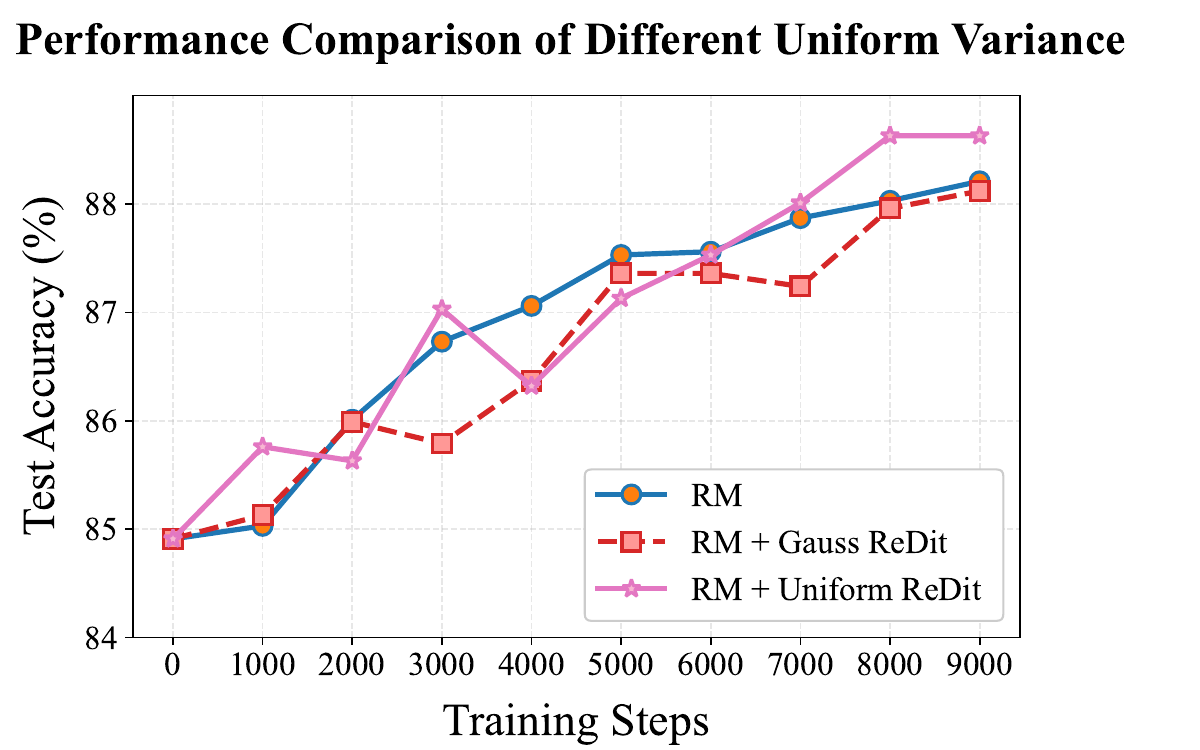}
        \caption{\ours{} has little effect on improving the performance of GRPO based RM.}
        \label{fig:rm-results}
    \end{minipage}
    \hfill
    \begin{minipage}{0.46\textwidth}
       \centering
        \includegraphics[width=1\textwidth]{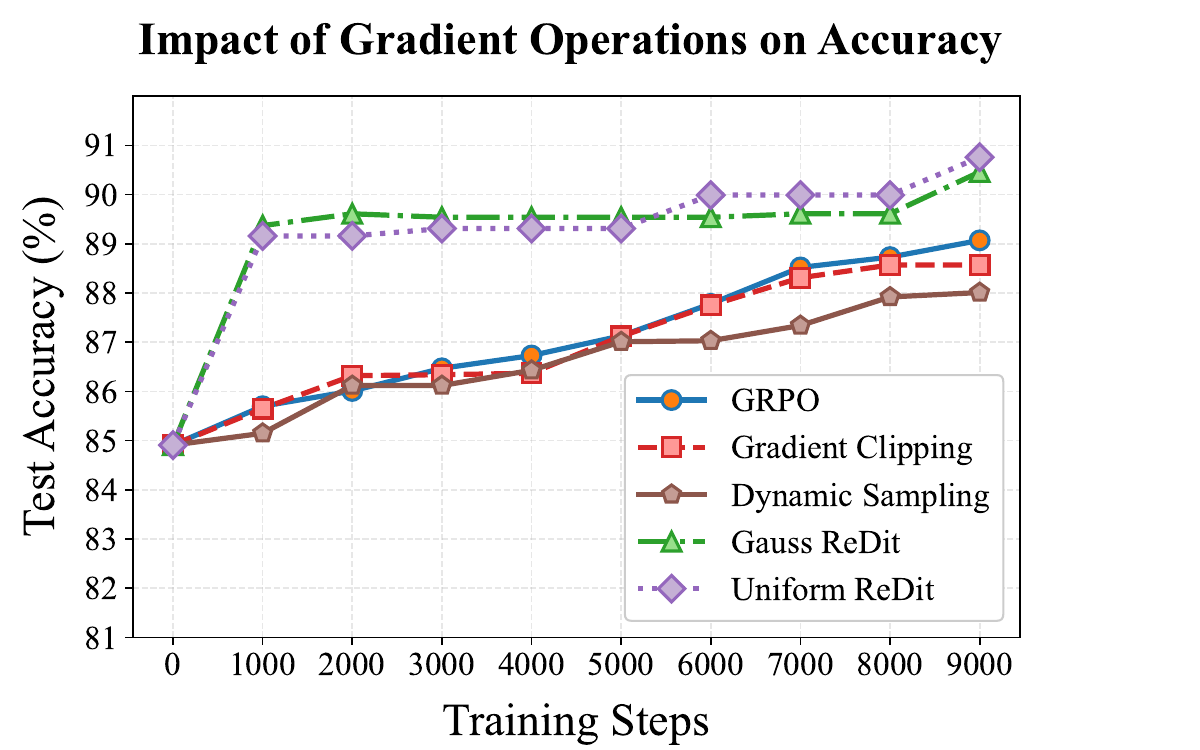}
        \caption{Appropriate perturbation achieves the best performance.}
        \label{fig:clip-results}
    \end{minipage}
\end{figure}

\textbf{Isolating the Effect on Discrete Rewards.}
To verify that the performance gains of \ours{} stem specifically from smoothing discrete rewards, we conducted a crucial ablation study. In this experiment, we replaced the discrete reward signal with a continuous one generated by a reward model pre-trained on human preference data. This model provides a continuous quality score within the range [0,1]. We then applied the \ours{} perturbation mechanism directly to these continuous rewards. The results, presented in Fig.~\ref{fig:rm-results}, show that applying \ours{} in this setting yields no discernible impact on either the convergence speed of model or its final performance. This outcome strongly indicates that the benefits of \ours{} are nullified when the reward landscape is already smooth. We therefore conclude that the efficacy of \ours{} lies specifically in addressing the optimization challenges inherent to sparse and discrete reward signals.

\textbf{Comparison with Direct Gradient Manipulation Baselines.}
We benchmark \ours{} against established techniques that directly address gradient instability: Gradient Clipping~\cite{Zhang}, which mitigates exploding gradients, and Dynamic Sampling~\cite{yu2025dapoopensourcellmreinforcement}, which alleviates vanishing gradients. The objective is to compare our \ours{} approach with methods that operate directly on the gradient signal. As illustrated in Figure~\ref{fig:clip-results}, \ours{} substantially outperforms both baseline methods. We attribute this performance gap to the inherent limitations of these heuristics. Gradient Clipping, for instance, crudely truncates gradient magnitudes, a non-principled operation that can introduce significant estimation bias. Conversely, while Dynamic Sampling can be effective for vanishing gradients, it offers no mechanism to prevent gradients from exploding. In contrast, \ours{} stabilizes the training process by smoothing the reward, which provides a more principled solution to prevent both gradient vanishing and explosion, thereby leading to more efficient and effective training.

\section{Theoretical Insights}
\label{sec:theoretical}

We provides a theoretical analyzing how perturbing discrete reward signals with, e.g., Gaussian noise, accelerates RL convergence, offering a principled explanation for observed empirical benefits.

\textbf{Problem Setup.}
Our analysis uses a simplified RL framework (from Eq.~\eqref{eq:rl_objective}) focusing on binary rewards $R(q,o) \in \{0,1\}$ for complete outputs (e.g., GRPO~\cite{shao2024deepseekmathpushinglimitsmathematical}), not token-level rewards. We investigate how Gaussian noise $\epsilon \sim \mathcal{N}(0,\sigma^2)$ injection improves convergence. The perturbed objective is:
\begin{equation}
\tilde{J}(\pi_\theta) = \mathbb{E}_{q \sim p_Q} \left[ \mathbb{E}_{o \sim \pi_\theta(\cdot|q)}\tilde{R}(q, o) \right],
\label{eq:noise_rl_objective}
\end{equation}
where the perturbed reward is $\tilde{R}(q, o) = R(q, o) + \epsilon$.

\begin{tcolorbox}[colback=blue!5!white,colframe=blue!70!white, left=0.5mm, right=1mm, top=1mm, bottom=1mm]
\begin{proposition}[Unbiased estimate of gradient]
\label{prop:unbiased}
\textit{\fontfamily{ppl}\selectfont
Introducing noise will still ensure the unbiased estimate of the gradient of the original optimization target Eq.~\eqref{eq:rl_objective}, that is:
\begin{equation}
\label{eq:mean}
\mathbb{E}\left[ \nabla_\theta \tilde{J}(\pi_\theta) \right] = \mathbb{E}\left[ \nabla_\theta J(\pi_\theta) \right]. \nonumber
\end{equation}}
\end{proposition}
\end{tcolorbox}
\textbf{Remark.} Proposition ~\ref{prop:unbiased} provides theoretical proof that introducing Gaussian noise perturbations into the discrete reward signal preserves the unbiased nature of the policy gradient estimate. This means that, under the perturbed reward, the expected direction of the policy update is consistent with the original objective being optimized. Maintaining this unbiased nature ensures that the injected noise does not introduce systematic biases into the learning dynamics, thus providing a theoretical basis for the empirical observation that our approach helps to consistently improve performance. See Appendix ~\ref{proof_pro1} for a detailed proof.

\begin{tcolorbox}[colback=blue!5!white,colframe=blue!70!white, left=0.5mm, right=1mm, top=1mm, bottom=1mm]
\begin{proposition}[Introducing the variance of gradient estimation]
\label{prop:variance}
\textit{\fontfamily{ppl}\selectfont
Suppose we are optimizing a non-degenerate strategy, that is, its gradient $\nabla_\theta \log \pi_\theta$ is not completely zero. Introducing noise will introduce gradient noise on the originally calculated gradient, and its variance is:
\begin{equation}
\label{eq:variance1}
\text{Var}(\text{Gradient Noise})= \sigma^2 \cdot \mathbb{E}\left[ \|\nabla_\theta \log \pi_\theta(o|q)\|^2 \right] > 0.  \nonumber
\end{equation}}
\end{proposition}
\end{tcolorbox}
\textbf{Remark.} In Proposition~\ref{prop:variance}, we analyze how Gaussian reward perturbations affect the variance of policy gradient estimates. Adding Gaussian noise \(\epsilon \sim \mathcal{N}(0, \sigma^2)\) to the reward introduces a "gradient noise" component proportional to \(\epsilon \cdot \nabla_\theta \log \pi_\theta(o|q)\) in the gradient estimate. The increased variance has significant optimization benefits: \textbf{Mitigate vanishing gradients:} Gradient noise provides consistent stochastic updates even when the original gradient terms are small or vanishing, thus helping to avoid flat regions. \textbf{Avoid exploding gradients:} The randomness induced by the noise enables the optimization trajectory to probabilistically bypass unstable regions of high curvature. Furthermore, the noise variance \(\sigma\) can be adjusted to control the magnitude of the gradient noise for optimal results. This mechanism enhances the robustness of policy optimization and explains the empirical improvements observed in training stability and convergence speed from reward perturbations. For detailed derivation, see Appendix~\ref{proof_pro2}.

\section{Limitations and Conclusions}
\label{sec:Conclusion}
\ours{} Improve the stability of reinforcement learning with zero-mean reward noise - theoretically smoothing gradients, preventing gradient instability, and accelerating convergence by increasing reward variance. Benchmarks verify significant improvements in convergence speed and performance. While our approach works, it requires careful tuning of the perturbation variance (although we adopt a small dataset search strategy, extensive experiments are still needed), and future work will aim to achieve automatic variance.

\newpage
\medskip
\bibliographystyle{unsrt}
\bibliography{neurips_2025}

\begin{thebibliography}{10}

\bibitem{AIAtMeta2024Llama}
{AI@Meta}.
\newblock The llama 4 herd.
\newblock \url{https://ai.meta.com/blog/llama-4-multimodal-intelligence/}, 2025.
\newblock Accessed: 2025.

\bibitem{Anthropic2024Claude}
{Anthropic}.
\newblock Claude 3.5 sonnet.
\newblock \url{https://www.anthropic.com/news/claude-3-5-sonnet}, 2024.
\newblock Accessed: 2024.

\bibitem{openai2024gpt4technicalreport}
OpenAI, Josh Achiam, Steven Adler, Sandhini Agarwal, Lama Ahmad, Ilge Akkaya, Florencia~Leoni Aleman, Diogo Almeida, Janko Altenschmidt, Sam Altman, Shyamal Anadkat, Red Avila, Igor Babuschkin, Suchir Balaji, Valerie Balcom, Paul Baltescu, Haiming Bao, Mohammad Bavarian, Jeff Belgum, Irwan Bello, Jake Berdine, Gabriel Bernadett-Shapiro, Christopher Berner, Lenny Bogdonoff, Oleg Boiko, Madelaine Boyd, Anna-Luisa Brakman, Greg Brockman, Tim Brooks, Miles Brundage, Kevin Button, Trevor Cai, Rosie Campbell, Andrew Cann, Brittany Carey, Chelsea Carlson, Rory Carmichael, Brooke Chan, Che Chang, Fotis Chantzis, Derek Chen, Sully Chen, Ruby Chen, Jason Chen, Mark Chen, Ben Chess, Chester Cho, Casey Chu, Hyung~Won Chung, Dave Cummings, Jeremiah Currier, Yunxing Dai, Cory Decareaux, Thomas Degry, Noah Deutsch, Damien Deville, Arka Dhar, David Dohan, Steve Dowling, Sheila Dunning, Adrien Ecoffet, Atty Eleti, Tyna Eloundou, David Farhi, Liam Fedus, Niko Felix, Simón~Posada Fishman, Juston Forte, Isabella Fulford, Leo
  Gao, Elie Georges, Christian Gibson, Vik Goel, Tarun Gogineni, Gabriel Goh, Rapha Gontijo-Lopes, Jonathan Gordon, Morgan Grafstein, Scott Gray, Ryan Greene, Joshua Gross, Shixiang~Shane Gu, Yufei Guo, Chris Hallacy, Jesse Han, Jeff Harris, Yuchen He, Mike Heaton, Johannes Heidecke, Chris Hesse, Alan Hickey, Wade Hickey, Peter Hoeschele, Brandon Houghton, Kenny Hsu, Shengli Hu, Xin Hu, Joost Huizinga, Shantanu Jain, Shawn Jain, Joanne Jang, Angela Jiang, Roger Jiang, Haozhun Jin, Denny Jin, Shino Jomoto, Billie Jonn, Heewoo Jun, Tomer Kaftan, Łukasz Kaiser, Ali Kamali, Ingmar Kanitscheider, Nitish~Shirish Keskar, Tabarak Khan, Logan Kilpatrick, Jong~Wook Kim, Christina Kim, Yongjik Kim, Jan~Hendrik Kirchner, Jamie Kiros, Matt Knight, Daniel Kokotajlo, Łukasz Kondraciuk, Andrew Kondrich, Aris Konstantinidis, Kyle Kosic, Gretchen Krueger, Vishal Kuo, Michael Lampe, Ikai Lan, Teddy Lee, Jan Leike, Jade Leung, Daniel Levy, Chak~Ming Li, Rachel Lim, Molly Lin, Stephanie Lin, Mateusz Litwin, Theresa Lopez, Ryan
  Lowe, Patricia Lue, Anna Makanju, Kim Malfacini, Sam Manning, Todor Markov, Yaniv Markovski, Bianca Martin, Katie Mayer, Andrew Mayne, Bob McGrew, Scott~Mayer McKinney, Christine McLeavey, Paul McMillan, Jake McNeil, David Medina, Aalok Mehta, Jacob Menick, Luke Metz, Andrey Mishchenko, Pamela Mishkin, Vinnie Monaco, Evan Morikawa, Daniel Mossing, Tong Mu, Mira Murati, Oleg Murk, David Mély, Ashvin Nair, Reiichiro Nakano, Rajeev Nayak, Arvind Neelakantan, Richard Ngo, Hyeonwoo Noh, Long Ouyang, Cullen O'Keefe, Jakub Pachocki, Alex Paino, Joe Palermo, Ashley Pantuliano, Giambattista Parascandolo, Joel Parish, Emy Parparita, Alex Passos, Mikhail Pavlov, Andrew Peng, Adam Perelman, Filipe de~Avila Belbute~Peres, Michael Petrov, Henrique~Ponde de~Oliveira~Pinto, Michael, Pokorny, Michelle Pokrass, Vitchyr~H. Pong, Tolly Powell, Alethea Power, Boris Power, Elizabeth Proehl, Raul Puri, Alec Radford, Jack Rae, Aditya Ramesh, Cameron Raymond, Francis Real, Kendra Rimbach, Carl Ross, Bob Rotsted, Henri Roussez,
  Nick Ryder, Mario Saltarelli, Ted Sanders, Shibani Santurkar, Girish Sastry, Heather Schmidt, David Schnurr, John Schulman, Daniel Selsam, Kyla Sheppard, Toki Sherbakov, Jessica Shieh, Sarah Shoker, Pranav Shyam, Szymon Sidor, Eric Sigler, Maddie Simens, Jordan Sitkin, Katarina Slama, Ian Sohl, Benjamin Sokolowsky, Yang Song, Natalie Staudacher, Felipe~Petroski Such, Natalie Summers, Ilya Sutskever, Jie Tang, Nikolas Tezak, Madeleine~B. Thompson, Phil Tillet, Amin Tootoonchian, Elizabeth Tseng, Preston Tuggle, Nick Turley, Jerry Tworek, Juan Felipe~Cerón Uribe, Andrea Vallone, Arun Vijayvergiya, Chelsea Voss, Carroll Wainwright, Justin~Jay Wang, Alvin Wang, Ben Wang, Jonathan Ward, Jason Wei, CJ~Weinmann, Akila Welihinda, Peter Welinder, Jiayi Weng, Lilian Weng, Matt Wiethoff, Dave Willner, Clemens Winter, Samuel Wolrich, Hannah Wong, Lauren Workman, Sherwin Wu, Jeff Wu, Michael Wu, Kai Xiao, Tao Xu, Sarah Yoo, Kevin Yu, Qiming Yuan, Wojciech Zaremba, Rowan Zellers, Chong Zhang, Marvin Zhang, Shengjia
  Zhao, Tianhao Zheng, Juntang Zhuang, William Zhuk, and Barret Zoph.
\newblock Gpt-4 technical report, 2024.

\bibitem{wei-etal-2025-flexora}
Chenxing Wei, Yao Shu, Ying~Tiffany He, and Fei Yu.
\newblock Flexora: Flexible low-rank adaptation for large language models.
\newblock In Wanxiang Che, Joyce Nabende, Ekaterina Shutova, and Mohammad~Taher Pilehvar, editors, {\em Proceedings of the 63rd Annual Meeting of the Association for Computational Linguistics (Volume 1: Long Papers)}, pages 14643--14682, Vienna, Austria, July 2025. Association for Computational Linguistics.

\bibitem{DeepReinforcementLearningFromHumanPreferences}
Paul~F. Christiano, Jan Leike, Tom~B. Brown, Miljan Martic, Shane Legg, and Dario Amodei.
\newblock Deep reinforcement learning from human preferences.
\newblock In {\em Proceedings of the 31st International Conference on Neural Information Processing Systems}, NIPS'17, page 4302–4310, Red Hook, NY, USA, 2017. Curran Associates Inc.

\bibitem{ziegler2019finetuning}
Daniel~M. Ziegler, Nisan Stiennon, Jeffrey Wu, Tom~B. Brown, Alec Radford, Dario Amodei, Paul Christiano, and Geoffrey Irving.
\newblock Fine-tuning language models from human preferences.
\newblock {\em arXiv preprint arXiv:1909.08593}, 2019.

\bibitem{lang-etal-2024-fine}
Hao Lang, Fei Huang, and Yongbin Li.
\newblock Fine-tuning language models with reward learning on policy.
\newblock In Kevin Duh, Helena Gomez, and Steven Bethard, editors, {\em Proceedings of the 2024 Conference of the North American Chapter of the Association for Computational Linguistics: Human Language Technologies (Volume 1: Long Papers)}, pages 1382--1392, Mexico City, Mexico, June 2024. Association for Computational Linguistics.

\bibitem{ouyang2022training}
Long Ouyang, Jeffrey Wu, Xu~Jiang, Diogo Almeida, Carroll Wainwright, Pamela Mishkin, Chong Zhang, Sandhini Agarwal, Katarina Slama, Alex Gray, John Schulman, Jacob Hilton, Fraser Kelton, Luke Miller, Maddie Simens, Amanda Askell, Peter Welinder, Paul Christiano, Jan Leike, and Ryan Lowe.
\newblock Training language models to follow instructions with human feedback.
\newblock In Alice~H. Oh, Alekh Agarwal, Danielle Belgrave, and Kyunghyun Cho, editors, {\em Advances in Neural Information Processing Systems}, 2022.

\bibitem{kaufmann2024surveyreinforcementlearninghuman}
Timo Kaufmann, Paul Weng, Viktor Bengs, and Eyke Hüllermeier.
\newblock A survey of reinforcement learning from human feedback, 2024.

\bibitem{lambert2025reinforcementlearninghumanfeedback}
Nathan Lambert.
\newblock Reinforcement learning from human feedback, 2025.

\bibitem{Cao_2024}
Yuji Cao, Huan Zhao, Yuheng Cheng, Ting Shu, Yue Chen, Guolong Liu, Gaoqi Liang, Junhua Zhao, Jinyue Yan, and Yun Li.
\newblock Survey on large language model-enhanced reinforcement learning: Concept, taxonomy, and methods.
\newblock {\em IEEE Transactions on Neural Networks and Learning Systems}, page 1–21, 2024.

\bibitem{rafailov2023direct}
Rafael Rafailov, Archit Sharma, Eric Mitchell, Christopher~D Manning, Stefano Ermon, and Chelsea Finn.
\newblock Direct preference optimization: Your language model is secretly a reward model.
\newblock In {\em Thirty-seventh Conference on Neural Information Processing Systems}, 2023.

\bibitem{wei2025testtimepolicyadaptationenhanced}
Chenxing Wei, Hong Wang, Ying He, Fei Yu, and Yao Shu.
\newblock Test-time policy adaptation for enhanced multi-turn interactions with llms, 2025.

\bibitem{deepseekai2025deepseekr1incentivizingreasoningcapability}
DeepSeek-AI, Daya Guo, Dejian Yang, Haowei Zhang, Junxiao Song, Ruoyu Zhang, Runxin Xu, Qihao Zhu, Shirong Ma, Peiyi Wang, Xiao Bi, Xiaokang Zhang, Xingkai Yu, Yu~Wu, Z.~F. Wu, Zhibin Gou, Zhihong Shao, Zhuoshu Li, Ziyi Gao, Aixin Liu, Bing Xue, Bingxuan Wang, Bochao Wu, Bei Feng, Chengda Lu, Chenggang Zhao, Chengqi Deng, Chenyu Zhang, Chong Ruan, Damai Dai, Deli Chen, Dongjie Ji, Erhang Li, Fangyun Lin, Fucong Dai, Fuli Luo, Guangbo Hao, Guanting Chen, Guowei Li, H.~Zhang, Han Bao, Hanwei Xu, Haocheng Wang, Honghui Ding, Huajian Xin, Huazuo Gao, Hui Qu, Hui Li, Jianzhong Guo, Jiashi Li, Jiawei Wang, Jingchang Chen, Jingyang Yuan, Junjie Qiu, Junlong Li, J.~L. Cai, Jiaqi Ni, Jian Liang, Jin Chen, Kai Dong, Kai Hu, Kaige Gao, Kang Guan, Kexin Huang, Kuai Yu, Lean Wang, Lecong Zhang, Liang Zhao, Litong Wang, Liyue Zhang, Lei Xu, Leyi Xia, Mingchuan Zhang, Minghua Zhang, Minghui Tang, Meng Li, Miaojun Wang, Mingming Li, Ning Tian, Panpan Huang, Peng Zhang, Qiancheng Wang, Qinyu Chen, Qiushi Du, Ruiqi Ge, Ruisong
  Zhang, Ruizhe Pan, Runji Wang, R.~J. Chen, R.~L. Jin, Ruyi Chen, Shanghao Lu, Shangyan Zhou, Shanhuang Chen, Shengfeng Ye, Shiyu Wang, Shuiping Yu, Shunfeng Zhou, Shuting Pan, S.~S. Li, Shuang Zhou, Shaoqing Wu, Shengfeng Ye, Tao Yun, Tian Pei, Tianyu Sun, T.~Wang, Wangding Zeng, Wanjia Zhao, Wen Liu, Wenfeng Liang, Wenjun Gao, Wenqin Yu, Wentao Zhang, W.~L. Xiao, Wei An, Xiaodong Liu, Xiaohan Wang, Xiaokang Chen, Xiaotao Nie, Xin Cheng, Xin Liu, Xin Xie, Xingchao Liu, Xinyu Yang, Xinyuan Li, Xuecheng Su, Xuheng Lin, X.~Q. Li, Xiangyue Jin, Xiaojin Shen, Xiaosha Chen, Xiaowen Sun, Xiaoxiang Wang, Xinnan Song, Xinyi Zhou, Xianzu Wang, Xinxia Shan, Y.~K. Li, Y.~Q. Wang, Y.~X. Wei, Yang Zhang, Yanhong Xu, Yao Li, Yao Zhao, Yaofeng Sun, Yaohui Wang, Yi~Yu, Yichao Zhang, Yifan Shi, Yiliang Xiong, Ying He, Yishi Piao, Yisong Wang, Yixuan Tan, Yiyang Ma, Yiyuan Liu, Yongqiang Guo, Yuan Ou, Yuduan Wang, Yue Gong, Yuheng Zou, Yujia He, Yunfan Xiong, Yuxiang Luo, Yuxiang You, Yuxuan Liu, Yuyang Zhou, Y.~X. Zhu,
  Yanhong Xu, Yanping Huang, Yaohui Li, Yi~Zheng, Yuchen Zhu, Yunxian Ma, Ying Tang, Yukun Zha, Yuting Yan, Z.~Z. Ren, Zehui Ren, Zhangli Sha, Zhe Fu, Zhean Xu, Zhenda Xie, Zhengyan Zhang, Zhewen Hao, Zhicheng Ma, Zhigang Yan, Zhiyu Wu, Zihui Gu, Zijia Zhu, Zijun Liu, Zilin Li, Ziwei Xie, Ziyang Song, Zizheng Pan, Zhen Huang, Zhipeng Xu, Zhongyu Zhang, and Zhen Zhang.
\newblock Deepseek-r1: Incentivizing reasoning capability in llms via reinforcement learning, 2025.

\bibitem{shao2024deepseekmathpushinglimitsmathematical}
Zhihong Shao, Peiyi Wang, Qihao Zhu, Runxin Xu, Junxiao Song, Xiao Bi, Haowei Zhang, Mingchuan Zhang, Y.~K. Li, Y.~Wu, and Daya Guo.
\newblock Deepseekmath: Pushing the limits of mathematical reasoning in open language models, 2024.

\bibitem{ActiveRewardLearning}
Dingwen Kong and Lin~F. Yang.
\newblock Provably feedback-efficient reinforcement learning via active reward learning.
\newblock In {\em Proceedings of the 36th International Conference on Neural Information Processing Systems}, NIPS '22, Red Hook, NY, USA, 2022. Curran Associates Inc.

\bibitem{wang2025crowdvlmr1expandingr1ability}
Zhiqiang Wang, Pengbin Feng, Yanbin Lin, Shuzhang Cai, Zongao Bian, Jinghua Yan, and Xingquan Zhu.
\newblock Crowdvlm-r1: Expanding r1 ability to vision language model for crowd counting using fuzzy group relative policy reward, 2025.

\bibitem{chan2023visionlanguage}
Harris Chan, Volodymyr Mnih, Feryal Behbahani, Michael Laskin, Luyu Wang, Fabio Pardo, Maxime Gazeau, Himanshu Sahni, Dan Horgan, Kate Baumli, Yannick Schroecker, Stephen Spencer, Richie Steigerwald, John Quan, Gheorghe Comanici, Sebastian Flennerhag, Alexander Neitz, Lei~M Zhang, Tom Schaul, Satinder Singh, Clare Lyle, Tim Rockt{\"a}schel, Jack Parker-Holder, and Kristian Holsheimer.
\newblock Vision-language models as a source of rewards.
\newblock In {\em Second Agent Learning in Open-Endedness Workshop}, 2023.

\bibitem{rengarajan2022reinforcement}
Desik Rengarajan, Gargi Vaidya, Akshay Sarvesh, Dileep Kalathil, and Srinivas Shakkottai.
\newblock Reinforcement learning with sparse rewards using guidance from offline demonstration.
\newblock In {\em International Conference on Learning Representations}, 2022.

\bibitem{vasan2024revisitingsparserewardsgoalreaching}
Gautham Vasan, Yan Wang, Fahim Shahriar, James Bergstra, Martin Jagersand, and A.~Rupam Mahmood.
\newblock Revisiting sparse rewards for goal-reaching reinforcement learning, 2024.

\bibitem{reward_shaping}
Prasoon Goyal, Scott Niekum, and Raymond~J. Mooney.
\newblock Using natural language for reward shaping in reinforcement learning.
\newblock In {\em Proceedings of the 28th International Joint Conference on Artificial Intelligence}, IJCAI'19, page 2385–2391. AAAI Press, 2019.

\bibitem{once-per-episodeFeedback}
Niladri~S. Chatterji, Aldo Pacchiano, Peter~L. Bartlett, and Michael~I. Jordan.
\newblock On the theory of reinforcement learning with once-per-episode feedback.
\newblock In {\em Proceedings of the 35th International Conference on Neural Information Processing Systems}, NIPS '21, Red Hook, NY, USA, 2021. Curran Associates Inc.

\bibitem{cao-etal-2024-enhancing}
Meng Cao, Lei Shu, Lei Yu, Yun Zhu, Nevan Wichers, Yinxiao Liu, and Lei Meng.
\newblock Enhancing reinforcement learning with dense rewards from language model critic.
\newblock In Yaser Al-Onaizan, Mohit Bansal, and Yun-Nung Chen, editors, {\em Proceedings of the 2024 Conference on Empirical Methods in Natural Language Processing}, pages 9119--9138, Miami, Florida, USA, November 2024. Association for Computational Linguistics.

\bibitem{chan2024dense}
Alex~James Chan, Hao Sun, Samuel Holt, and Mihaela van~der Schaar.
\newblock Dense reward for free in reinforcement learning from human feedback.
\newblock In {\em International Conference on Machine Learning}, 2024.

\bibitem{xie2024textreward}
Tianbao Xie, Siheng Zhao, Chen~Henry Wu, Yitao Liu, Qian Luo, Victor Zhong, Yanchao Yang, and Tao Yu.
\newblock Text2reward: Reward shaping with language models for reinforcement learning.
\newblock In {\em The Twelfth International Conference on Learning Representations}, 2024.

\bibitem{gradient-basedReinforcementLearning}
Lex Weaver and Nigel Tao.
\newblock The optimal reward baseline for gradient-based reinforcement learning.
\newblock In {\em Proceedings of the Seventeenth Conference on Uncertainty in Artificial Intelligence}, UAI'01, page 538–545, San Francisco, CA, USA, 2001. Morgan Kaufmann Publishers Inc.

\bibitem{razin2024vanishing}
Noam Razin, Hattie Zhou, Omid Saremi, Vimal Thilak, Arwen Bradley, Preetum Nakkiran, Joshua~M. Susskind, and Etai Littwin.
\newblock Vanishing gradients in reinforcement finetuning of language models.
\newblock In {\em The Twelfth International Conference on Learning Representations}, 2024.

\bibitem{GradientMonitoredReinforcementLearning}
Mohammed~Sharafath Abdul~Hameed, Gavneet~Singh Chadha, Andreas Schwung, and Steven~X. Ding.
\newblock Gradient monitored reinforcement learning.
\newblock {\em IEEE Transactions on Neural Networks and Learning Systems}, 34(8):4106--4119, 2023.

\bibitem{Evolution-guided}
Shauharda Khadka and Kagan Tumer.
\newblock Evolution-guided policy gradient in reinforcement learning.
\newblock In {\em Proceedings of the 32nd International Conference on Neural Information Processing Systems}, NIPS'18, page 1196–1208, Red Hook, NY, USA, 2018. Curran Associates Inc.

\bibitem{zhang2025gvpogroupvariancepolicy}
Kaichen Zhang, Yuzhong Hong, Junwei Bao, Hongfei Jiang, Yang Song, Dingqian Hong, and Hui Xiong.
\newblock Gvpo: Group variance policy optimization for large language model post-training, 2025.

\bibitem{qwen2025qwen25technicalreport}
Qwen, :, An~Yang, Baosong Yang, Beichen Zhang, Binyuan Hui, Bo~Zheng, Bowen Yu, Chengyuan Li, Dayiheng Liu, Fei Huang, Haoran Wei, Huan Lin, Jian Yang, Jianhong Tu, Jianwei Zhang, Jianxin Yang, Jiaxi Yang, Jingren Zhou, Junyang Lin, Kai Dang, Keming Lu, Keqin Bao, Kexin Yang, Le~Yu, Mei Li, Mingfeng Xue, Pei Zhang, Qin Zhu, Rui Men, Runji Lin, Tianhao Li, Tianyi Tang, Tingyu Xia, Xingzhang Ren, Xuancheng Ren, Yang Fan, Yang Su, Yichang Zhang, Yu~Wan, Yuqiong Liu, Zeyu Cui, Zhenru Zhang, and Zihan Qiu.
\newblock Qwen2.5 technical report, 2025.

\bibitem{cobbe2021trainingverifierssolvemath}
Karl Cobbe, Vineet Kosaraju, Mohammad Bavarian, Mark Chen, Heewoo Jun, Lukasz Kaiser, Matthias Plappert, Jerry Tworek, Jacob Hilton, Reiichiro Nakano, Christopher Hesse, and John Schulman.
\newblock Training verifiers to solve math word problems, 2021.

\bibitem{Zhang2020Why}
Jingzhao Zhang, Tianxing He, Suvrit Sra, and Ali Jadbabaie.
\newblock Why gradient clipping accelerates training: A theoretical justification for adaptivity.
\newblock In {\em International Conference on Learning Representations}, 2020.

\bibitem{ivison2024unpacking}
Hamish Ivison, Yizhong Wang, Jiacheng Liu, Zeqiu Wu, Valentina Pyatkin, Nathan Lambert, Noah~A. Smith, Yejin Choi, and Hannaneh Hajishirzi.
\newblock Unpacking {DPO} and {PPO}: Disentangling best practices for learning from preference feedback.
\newblock In {\em The Thirty-eighth Annual Conference on Neural Information Processing Systems}, 2024.

\bibitem{chen-etal-2024-accuracy}
Yanjun Chen, Dawei Zhu, Yirong Sun, Xinghao Chen, Wei Zhang, and Xiaoyu Shen.
\newblock The accuracy paradox in {RLHF}: When better reward models don`t yield better language models.
\newblock In Yaser Al-Onaizan, Mohit Bansal, and Yun-Nung Chen, editors, {\em Proceedings of the 2024 Conference on Empirical Methods in Natural Language Processing}, pages 2980--2989, Miami, Florida, USA, November 2024. Association for Computational Linguistics.

\bibitem{wen2025rethinking}
Xueru Wen, Jie Lou, Yaojie Lu, Hongyu Lin, XingYu, Xinyu Lu, Ben He, Xianpei Han, Debing Zhang, and Le~Sun.
\newblock Rethinking reward model evaluation: Are we barking up the wrong tree?
\newblock In {\em The Thirteenth International Conference on Learning Representations}, 2025.

\bibitem{hallak2015contextualmarkovdecisionprocesses}
Assaf Hallak, Dotan~Di Castro, and Shie Mannor.
\newblock Contextual markov decision processes, 2015.

\bibitem{yu2025dapoopensourcellmreinforcement}
Qiying Yu, Zheng Zhang, Ruofei Zhu, Yufeng Yuan, Xiaochen Zuo, Yu~Yue, Tiantian Fan, Gaohong Liu, Lingjun Liu, Xin Liu, Haibin Lin, Zhiqi Lin, Bole Ma, Guangming Sheng, Yuxuan Tong, Chi Zhang, Mofan Zhang, Wang Zhang, Hang Zhu, Jinhua Zhu, Jiaze Chen, Jiangjie Chen, Chengyi Wang, Hongli Yu, Weinan Dai, Yuxuan Song, Xiangpeng Wei, Hao Zhou, Jingjing Liu, Wei-Ying Ma, Ya-Qin Zhang, Lin Yan, Mu~Qiao, Yonghui Wu, and Mingxuan Wang.
\newblock Dapo: An open-source llm reinforcement learning system at scale, 2025.

\bibitem{liu2025understandingr1zeroliketrainingcritical}
Zichen Liu, Changyu Chen, Wenjun Li, Penghui Qi, Tianyu Pang, Chao Du, Wee~Sun Lee, and Min Lin.
\newblock Understanding r1-zero-like training: A critical perspective, 2025.

\bibitem{hu2025reinforceefficientrlhfalgorithm}
Jian Hu, Jason~Klein Liu, and Wei Shen.
\newblock Reinforce++: An efficient rlhf algorithm with robustness to both prompt and reward models, 2025.

\bibitem{razin2025makesrewardmodelgood}
Noam Razin, Zixuan Wang, Hubert Strauss, Stanley Wei, Jason~D. Lee, and Sanjeev Arora.
\newblock What makes a reward model a good teacher? an optimization perspective, 2025.

\bibitem{hendrycks2021measuringmathematicalproblemsolving}
Dan Hendrycks, Collin Burns, Saurav Kadavath, Akul Arora, Steven Basart, Eric Tang, Dawn Song, and Jacob Steinhardt.
\newblock Measuring mathematical problem solving with the math dataset, 2021.

\bibitem{lu2021intergpsinterpretablegeometryproblem}
Pan Lu, Ran Gong, Shibiao Jiang, Liang Qiu, Siyuan Huang, Xiaodan Liang, and Song-Chun Zhu.
\newblock Inter-gps: Interpretable geometry problem solving with formal language and symbolic reasoning, 2021.

\bibitem{shao2025spuriousrewardsrethinkingtraining}
Rulin Shao, Shuyue~Stella Li, Rui Xin, Scott Geng, Yiping Wang, Sewoong Oh, Simon~Shaolei Du, Nathan Lambert, Sewon Min, Ranjay Krishna, Yulia Tsvetkov, Hannaneh Hajishirzi, Pang~Wei Koh, and Luke Zettlemoyer.
\newblock Spurious rewards: Rethinking training signals in rlvr, 2025.

\bibitem{jiang2023mistral7b}
Albert~Q. Jiang, Alexandre Sablayrolles, Arthur Mensch, Chris Bamford, Devendra~Singh Chaplot, Diego de~las Casas, Florian Bressand, Gianna Lengyel, Guillaume Lample, Lucile Saulnier, Lélio~Renard Lavaud, Marie-Anne Lachaux, Pierre Stock, Teven~Le Scao, Thibaut Lavril, Thomas Wang, Timothée Lacroix, and William~El Sayed.
\newblock Mistral 7b, 2023.

\bibitem{hu2022lora}
Edward~J Hu, Yelong Shen, Phillip Wallis, Zeyuan Allen-Zhu, Yuanzhi Li, Shean Wang, Lu~Wang, and Weizhu Chen.
\newblock Lo{RA}: Low-rank adaptation of large language models.
\newblock In {\em International Conference on Learning Representations}, 2022.

\bibitem{vonwerra2022trl}
Leandro von Werra, Younes Belkada, Lewis Tunstall, Edward Beeching, Tristan Thrush, Nathan Lambert, Shengyi Huang, Kashif Rasul, and Quentin Gallouédec.
\newblock Trl: Transformer reinforcement learning.
\newblock \url{https://github.com/huggingface/trl}, 2020.

\bibitem{2023opencompass}
OpenCompass Contributors.
\newblock Opencompass: A universal evaluation platform for foundation models.
\newblock \url{https://github.com/open-compass/opencompass}, 2023.

\bibitem{Zhang}
Jingzhao Zhang, Tianxing He, Suvrit Sra, and Ali Jadbabaie.
\newblock Why gradient clipping accelerates training: A theoretical justification for adaptivity.
\newblock In {\em International Conference on Learning Representations}, 2020.

\bibitem{zhang2024cppo}
Han Zhang, Yu~Lei, Lin Gui, Min Yang, Yulan He, Hui Wang, and Ruifeng Xu.
\newblock {CPPO}: Continual learning for reinforcement learning with human feedback.
\newblock In {\em The Twelfth International Conference on Learning Representations}, 2024.

\bibitem{chu2025gpgsimplestrongreinforcement}
Xiangxiang Chu, Hailang Huang, Xiao Zhang, Fei Wei, and Yong Wang.
\newblock Gpg: A simple and strong reinforcement learning baseline for model reasoning, 2025.

\bibitem{Policy_Invariance}
Andrew~Y. Ng, Daishi Harada, and Stuart~J. Russell.
\newblock Policy invariance under reward transformations: Theory and application to reward shaping.
\newblock In {\em Proceedings of the Sixteenth International Conference on Machine Learning}, ICML '99, page 278–287, San Francisco, CA, USA, 1999. Morgan Kaufmann Publishers Inc.

\bibitem{reward_shaping_in_rl}
Marek Grze\'{s}.
\newblock Reward shaping in episodic reinforcement learning.
\newblock In {\em Proceedings of the 16th Conference on Autonomous Agents and MultiAgent Systems}, AAMAS '17, page 565–573, Richland, SC, 2017. International Foundation for Autonomous Agents and Multiagent Systems.

\bibitem{gupta2022unpacking}
Abhishek Gupta, Aldo Pacchiano, Yuexiang Zhai, Sham~M. Kakade, and Sergey Levine.
\newblock Unpacking reward shaping: Understanding the benefits of reward engineering on sample complexity.
\newblock In Alice~H. Oh, Alekh Agarwal, Danielle Belgrave, and Kyunghyun Cho, editors, {\em Advances in Neural Information Processing Systems}, 2022.

\bibitem{fazel2018global}
Maryam Fazel, Rong Ge, Sham~M. Kakade, and Mehran Mesbahi.
\newblock Global convergence of policy gradient methods for linearized control problems, 2018.

\bibitem{wei2025paftpromptagnosticfinetuning}
Chenxing Wei, Yao Shu, Mingwen Ou, Ying~Tiffany He, and Fei~Richard Yu.
\newblock Paft: Prompt-agnostic fine-tuning, 2025.

\end{thebibliography}
\newpage

\section*{Acknowledgement}
This work was supported by the Shenzhen Science and Technology Program (Grant No. ZDSYS20220527171400002), the National Natural Science Foundation of China (Grant Nos. 62271324, 62231020, and 62371309), and the Open Research Fund from Guangdong Laboratory of Artificial Intelligence and Digital Economy (SZ) (Grant No. GML-KF-24-32).

\section*{NeurIPS Paper Checklist}

The checklist is designed to encourage best practices for responsible machine learning research, addressing issues of reproducibility, transparency, research ethics, and societal impact. Do not remove the checklist: {\bf The papers not including the checklist will be desk rejected.} The checklist should follow the references and follow the (optional) supplemental material.  The checklist does NOT count towards the page
limit. 

Please read the checklist guidelines carefully for information on how to answer these questions. For each question in the checklist:
\begin{itemize}
    \item You should answer \answerYes{}, \answerNo{}, or \answerNA{}.
    \item \answerNA{} means either that the question is Not Applicable for that particular paper or the relevant information is Not Available.
    \item Please provide a short (1–2 sentence) justification right after your answer (even for NA). 
\end{itemize}

{\bf The checklist answers are an integral part of your paper submission.} They are visible to the reviewers, area chairs, senior area chairs, and ethics reviewers. You will be asked to also include it (after eventual revisions) with the final version of your paper, and its final version will be published with the paper.

The reviewers of your paper will be asked to use the checklist as one of the factors in their evaluation. While "\answerYes{}" is generally preferable to "\answerNo{}", it is perfectly acceptable to answer "\answerNo{}" provided a proper justification is given (e.g., "error bars are not reported because it would be too computationally expensive" or "we were unable to find the license for the dataset we used"). In general, answering "\answerNo{}" or "\answerNA{}" is not grounds for rejection. While the questions are phrased in a binary way, we acknowledge that the true answer is often more nuanced, so please just use your best judgment and write a justification to elaborate. All supporting evidence can appear either in the main paper or the supplemental material, provided in appendix. If you answer \answerYes{} to a question, in the justification please point to the section(s) where related material for the question can be found.

IMPORTANT, please:
\begin{itemize}
    \item {\bf Delete this instruction block, but keep the section heading ``NeurIPS Paper Checklist"},
    \item  {\bf Keep the checklist subsection headings, questions/answers and guidelines below.}
    \item {\bf Do not modify the questions and only use the provided macros for your answers}.
\end{itemize}


\begin{enumerate}

\item {\bf Claims}
    \item[] Question: Do the main claims made in the abstract and introduction accurately reflect the paper's contributions and scope?
    \item[] Answer: \answerYes{} 
    \item[] Justification: The claims are put in the abstract and Section~\ref{introduction}

\item {\bf Limitations}
    \item[] Question: Does the paper discuss the limitations of the work performed by the authors?
    \item[] Answer: \answerYes{} 
    \item[] Justification: For further details, please refer to the "Limitations and Conclusions" section, found in Section~\ref{sec:Conclusion}.

\item {\bf Theory assumptions and proofs}
    \item[] Question: For each theoretical result, does the paper provide the full set of assumptions and a complete (and correct) proof?
    \item[] Answer: \answerYes{} 
    \item[] Justification: The claims are put in the abstract and Appendix~\ref{sec:math}.

\item {\bf Experimental result reproducibility}
    \item[] Question: Does the paper fully disclose all the information needed to reproduce the main experimental results of the paper to the extent that it affects the main claims and/or conclusions of the paper (regardless of whether the code and data are provided or not)?
    \item[] Answer: \answerYes{} 
    \item[] Justification: For detailed information, please refer to the "Method" section in Section~\ref{sec:ours_method}. Complete pseudocode can be found in Algorithm~\ref{alg:reward_smoothing}. Specific experimental settings are discussed in  Appendix~\ref{sec:setting}.

\item {\bf Open access to data and code}
    \item[] Question: Does the paper provide open access to the data and code, with sufficient instructions to faithfully reproduce the main experimental results, as described in supplemental material?
    \item[] Answer: \answerYes{} 
    \item[] Justification: We provide the instruction and code in supplemental material.

\item {\bf Experimental setting/details}
    \item[] Question: Does the paper specify all the training and test details (e.g., data splits, hyperparameters, how they were chosen, type of optimizer, etc.) necessary to understand the results?
    \item[] Answer: \answerYes{} 
    \item[] Justification: Specific experimental settings are discussed in  Appendix~\ref{sec:setting}.

\item {\bf Experiment statistical significance}
    \item[] Question: Does the paper report error bars suitably and correctly defined or other appropriate information about the statistical significance of the experiments?
    \item[] Answer: \answerYes{} 
    \item[] Justification:  For a detailed comparison, please refer to the experimental sections in Section~\ref{sec:results}.

\item {\bf Experiments compute resources}
    \item[] Question: For each experiment, does the paper provide sufficient information on the computer resources (type of compute workers, memory, time of execution) needed to reproduce the experiments?
    \item[] Answer: \answerNo{} 
    \item[] Justification: Specific experimental environments can be found in Sec~\ref{sec:experimental settings}.
    
\item {\bf Code of ethics}
    \item[] Question: Does the research conducted in the paper conform, in every respect, with the NeurIPS Code of Ethics \url{https://neurips.cc/public/EthicsGuidelines}?
    \item[] Answer: \answerYes{} 
    \item[] Justification: We have checked the NeurIPS code of ethics.

\item {\bf Broader impacts}
    \item[] Question: Does the paper discuss both potential positive societal impacts and negative societal impacts of the work performed?
    \item[] Answer: \answerNA{} 
    \item[] Justification: This paper mainly studies the impact of discrete rewards on reinforcement learning. It does not have any direct positive or negative social impacts.
    
\item {\bf Safeguards}
    \item[] Question: Does the paper describe safeguards that have been put in place for responsible release of data or models that have a high risk for misuse (e.g., pretrained language models, image generators, or scraped datasets)?
    \item[] Answer: \answerNA{} 
    \item[] Justification: We have checked this and confirmed the paper poses no such risks.

\item {\bf Licenses for existing assets}
    \item[] Question: Are the creators or original owners of assets (e.g., code, data, models), used in the paper, properly credited and are the license and terms of use explicitly mentioned and properly respected?
    \item[] Answer: \answerYes{} 
    \item[] Justification: The original papers or URLs of the codes, models, and data sets used in this article have been cited in the paper.

\item {\bf New assets}
    \item[] Question: Are new assets introduced in the paper well documented and is the documentation provided alongside the assets?
    \item[] Answer: \answerYes{} 
    \item[] Justification: We provide the instruction in supplemental material about our code and data.

\item {\bf Crowdsourcing and research with human subjects}
    \item[] Question: For crowdsourcing experiments and research with human subjects, does the paper include the full text of instructions given to participants and screenshots, if applicable, as well as details about compensation (if any)? 
    \item[] Answer: \answerNA{} 
    \item[] Justification: The paper does not involve crowdsourcing nor research with human subjects.

\item {\bf Institutional review board (IRB) approvals or equivalent for research with human subjects}
    \item[] Question: Does the paper describe potential risks incurred by study participants, whether such risks were disclosed to the subjects, and whether Institutional Review Board (IRB) approvals (or an equivalent approval/review based on the requirements of your country or institution) were obtained?
    \item[] Answer: \answerNA{} 
    \item[] Justification: The paper does not involve crowdsourcing nor research with human subjects.

\item {\bf Declaration of LLM usage}
    \item[] Question: Does the paper describe the usage of LLMs if it is an important, original, or non-standard component of the core methods in this research? Note that if the LLM is used only for writing, editing, or formatting purposes and does not impact the core methodology, scientific rigorousness, or originality of the research, declaration is not required.
    \item[] Answer: \answerNA{} 
    \item[] Justification: This paper only utilizes LLMs for writing and editing.

\end{enumerate}

\newpage
\appendix
\section{Related Work}
\label{sec:related_work}
\textbf{Reinforcement Learning with Discrete Rewards.}  
Group Relative Policy Optimization (GRPO)~\cite{shao2024deepseekmathpushinglimitsmathematical} utilizes discrete rewards generated by a rule-based reward function to guide the policy model update. This reward function, known for its simplicity and unbiasedness, effectively mitigates reward hacking and has demonstrated strong performance. However, GRPO faces challenges related to slow training speed and unstable gradients during training. To address these issues, various methods have been proposed. DAPO~\cite{yu2025dapoopensourcellmreinforcement} introduced a dynamic sampling strategy to improve gradient effectiveness by dynamically filtering invalid samples, thereby increasing sample efficiency, although this reduced training speed. CPPO~\cite{zhang2024cppo} prunes completions with low absolute advantages, significantly reducing the number of gradient calculations and updates required, which enhances training efficiency but can lead to gradient estimation errors. GPG~\cite{chu2025gpgsimplestrongreinforcement} directly optimizes the original reinforcement learning objective, eliminating the need for a proxy loss function and improving training efficiency. However, this simplification may result in a significant divergence between the actor and policy models. Dr.GRPO~\cite{liu2025understandingr1zeroliketrainingcritical} improves token efficiency while maintaining inference performance. Despite these efforts, a critical challenge remains: these algorithms largely neglect the inherent difficulties introduced by discrete rewards during the optimization process. The oscillations caused by gradient vanishing and exploding are major contributors to the slow optimization speed. Our work specifically aims to overcome the challenges in gradient optimization that arise from using discrete rewards.

\textbf{Addressing Reward Design Challenges in LLM Reinforcement Learning.}
Designing effective reward functions for identifying optimal strategies is a well-established area of research outside the context of Large Language Models (LLMs)~\cite{Policy_Invariance, reward_shaping_in_rl, gupta2022unpacking}. However, a consensus on the optimal approach for LLM reinforcement learning has not yet been reached~\cite{fazel2018global}. The RLHF framework proposed training a reward model to score LLM outputs~\cite{DeepReinforcementLearningFromHumanPreferences}. A recurring challenge, as noted by numerous studies, is that low reward model accuracy can induce reward hacking~\cite{ivison2024unpacking, chen-etal-2024-accuracy, wen2025rethinking}. Conversely, improving accuracy often reduces reward variance, which can slow down policy model convergence due to vanishing gradients~\cite{razin2024vanishing}. Although the reward function presented in GRPO provides perfectly correct rewards, thereby avoiding reward hacking, it exacerbates gradient instability and hinders optimization speed~\cite{shao2024deepseekmathpushinglimitsmathematical, liu2025understandingr1zeroliketrainingcritical}. Recent theoretical findings indicate that a successful reward function requires a trade-off between variance and inaccuracy~\cite{razin2025makesrewardmodelgood}. Motivated by this, our work seeks to design a reward function that effectively addresses the problem of reward hacking while simultaneously facilitating efficient optimization.

\section{Theorems and proofs} \label{sec:math}

\subsection{Definitions}\label{definition}
From Definition 1, 2 in \cite{razin2025makesrewardmodelgood}, The accuracy and variance of the reward function is as follows:

\begin{definition} \label{accuracy_definition}
Given a prompt $x \in \mathcal{X}$, the accuracy of a reward model $r_{RM}: \mathcal{X} \times \mathcal{Y} \to [-1, 1]$ with respect to a distribution $\mathcal{D}$ over unordered output pairs is defined by:

\begin{equation}
\text{Acc}_{x,\mathcal{D}}(r_{RM}) := \mathbb{E}_{\{y,y'\} \sim \mathcal{D}} \Biggl[ \mathbbm{1} \biggl[ \text{sign}\bigl(r_{RM}(x,y) - r_{RM}(x,y')\bigr) = \text{sign}\bigl(r_G(x,y) - r_G(x,y')\bigr) \biggr] \Biggr],
\end{equation}

where $r_G$ is the ground truth reward, $\mathbbm{1}[\cdot]$ is an indicator function, and $\text{sign}: \mathbb{R} \to \{-1, 0, 1\}$ is the sign function.\footnote{For a set of prompts, accuracy refers to the mean accuracy over the set.}
\end{definition}

\begin{definition} \label{variance_definition}
Given a policy $\pi_\theta$, prompt $x \in \mathcal{X}$, and reward model $r_{RM}: \mathcal{X} \times \mathcal{Y} \to [-1, 1]$, the reward variance induced by $r_{RM}$ for $\pi_\theta$ and $x$ is defined by:
\begin{equation}
\text{Var}_{y \sim \pi_\theta(\cdot|x)}[r_{RM}(x, y)] := \mathbb{E}_{y \sim \pi_\theta(\cdot|x)} \Biggl[ \biggl( r_{RM}(x, y) - \mathbb{E}_{y' \sim \pi_\theta(\cdot|x)}\bigl[r_{RM}(x, y')\bigr] \biggr)^2 \Biggr].
\end{equation}
\end{definition}

\subsection{Proof of Proposition~\ref{prop:unbiased}}\label{proof_pro1}
The proof of Proposition~\ref{prop:unbiased} is expressed as follows:
\begin{proof}
By the policy gradient theorem, the gradient of the original objective~\eqref{eq:rl_objective} expands to:
\begin{equation}
\nabla_\theta J(\pi_\theta) = \mathbb{E}_{q \sim p_Q} \mathbb{E}_{o \sim \pi_\theta(\cdot|q)} \left[ R(q,o) \nabla_\theta \log \pi_\theta(o|q) \right].
\end{equation}

For the noise-injected objective, its gradient becomes:
\begin{equation}
\nabla_\theta \tilde{J}(\pi_\theta) = \mathbb{E}_{q \sim p_Q} \mathbb{E}_{o \sim \pi_\theta(\cdot|q)} \left[ \tilde{R}(q,o) \nabla_\theta \log \pi_\theta(o|q) \right].
\end{equation}

Substituting $\tilde{R}(q,o) = R(q,o) + \epsilon$ and leveraging linearity of expectation:
\begin{align}
\mathbb{E}\left[ \nabla_\theta \tilde{J} \right] &= \mathbb{E}_{q,o,\epsilon} \left[ (R(q,o)+\epsilon) \nabla_\theta \log \pi_\theta(o|q) \right] \\
&= \underbrace{\mathbb{E}_{q,o} \left[ R(q,o) \nabla_\theta \log \pi_\theta(o|q) \right]}_{\mathbb{E}[\nabla_\theta J]} 
+ \mathbb{E}_{\epsilon}[\epsilon] \cdot \mathbb{E}_{q,o} \left[ \nabla_\theta \log \pi_\theta(o|q) \right].
\end{align}
Zero-mean noise: $\mathbb{E}_\epsilon[\epsilon] = 0$ by definition of $\mathcal{N}(0,\sigma^2)$.
Thus, the cross-term vanishes:
\begin{equation}
\mathbb{E}[ \nabla_\theta \tilde{J} ] = \mathbb{E}[\nabla_\theta J] + 0 = \mathbb{E}[\nabla_\theta J].
\end{equation}
\end{proof}

\subsection{Proof of Proposition~\ref{prop:variance}    }\label{proof_pro2}
\begin{proof}

Consider the perturbed objective function with noise-augmented reward $R(q,o) + \epsilon$. The estimated value of the gradient of the noise enhancement objective function using $n$ samples is:
\begin{equation}
\nabla \hat{\tilde{J}}(\theta) = \frac{1}{n}\sum_{i=1}^{n}\left[ \nabla \log \pi_\theta(o_i|q_i) \cdot \left( R(q_i,o_i) + \epsilon_i \right) \right], \label{eq:perturbed_grad}
\end{equation}

where $\epsilon_i \sim \mathcal{N}(0, \sigma^2)$ is the Gaussian noise.
The original reward gradient is:
\begin{equation}
\nabla \hat{J}(\theta) = \frac{1}{n}\sum_{i=1}^{n}\left[ \nabla \log \pi_\theta(o_i|q_i) \cdot \left( R(q_i,o_i) \right) \right]. 
\label{eq:flat_region}
\end{equation}

Under this condition, the Eq.~\eqref{eq:perturbed_grad} simplifies to:
\begin{equation}
\nabla \hat{\tilde{J}}(\theta) = \underbrace{\nabla \hat{J}(\theta)}_{\text{origin gradient}} + \underbrace{\frac{1}{n}\sum_{i=1}^{n}\left[ \nabla \log \pi_\theta(o_i|q_i) \cdot \epsilon_i \right]}_{\text{noise gradient}}. \label{eq:noise_component}
\end{equation}

While the expectation $\mathbb{E}_\epsilon[\epsilon] = 0$ implies the noise contribution's mean is zero, the variance of the gradient term persists. To compute this variance, we use the definition: $\text{Var}(X) = \mathbb{E}[X^2] - (\mathbb{E}[X])^2$.
Applying this to the noise-induced component $\epsilon \cdot \nabla_\theta \log \pi_\theta(o|q)$, we get:
\begin{equation}
\text{Var}\left( \epsilon \cdot \nabla_\theta \log \pi_\theta(o|q) \right) = \mathbb{E}\left[ \epsilon^2 \cdot \|\nabla_\theta \log \pi_\theta(o|q)\|^2 \right] - \left( \mathbb{E}\left[ \epsilon \cdot \nabla_\theta \log \pi_\theta(o|q) \right] \right)^2. \label{eq:variance_step}
\end{equation}

Since $\mathbb{E}[\epsilon] = 0$, the second term vanishes. For the first term, note that:
\begin{equation}
\mathbb{E}[\epsilon^2] = \text{Var}(\epsilon) + (\mathbb{E}[\epsilon])^2 = \sigma^2 + 0 = \sigma^2.
\end{equation}
This allows us to simplify the variance expression to:
\begin{equation}
\text{Var(noise gradient)} = \text{Var}\left( \epsilon \cdot \nabla_\theta \log \pi_\theta \right) = \sigma^2 \cdot \mathbb{E}\left[ \|\nabla_\theta \log \pi_\theta(o|q)\|^2 \right] > 0, \label{eq:variance}
\end{equation}
provided $\nabla_\theta \log \pi_\theta$ is not identically zero (a reasonable assumption for non-degenerate policies).

\end{proof}

\section{Training Dynamic}
In this section, we show more Training Dynamic information.

\begin{figure*}[ht]
\centering
\includegraphics[width=1.0\textwidth]{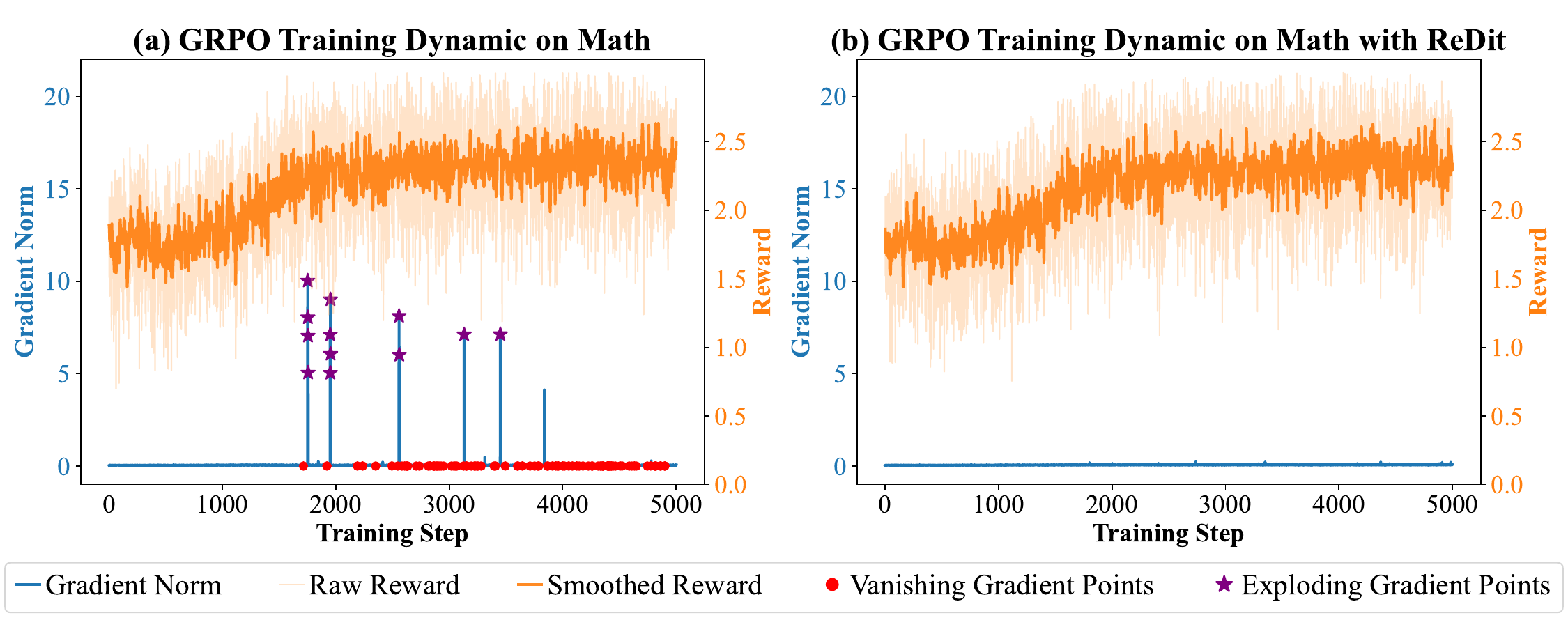}
\vspace{-5mm}
\caption{Training Dynamics of Gradient Norm and Reward on Math Dataset.}
\label{fig:Training Dynamics_1}
\vspace{-3mm}
\end{figure*}
Figure ~\ref{fig:Training Dynamics_1} shows the training dynamics of using and not using \ours{} on the Math dataset, indicating that using \ours{} can solve the problems of gradient oscillation and gradient vanishing, and improve training stability

\begin{figure*}[ht]
\centering
\includegraphics[width=1.0\textwidth]{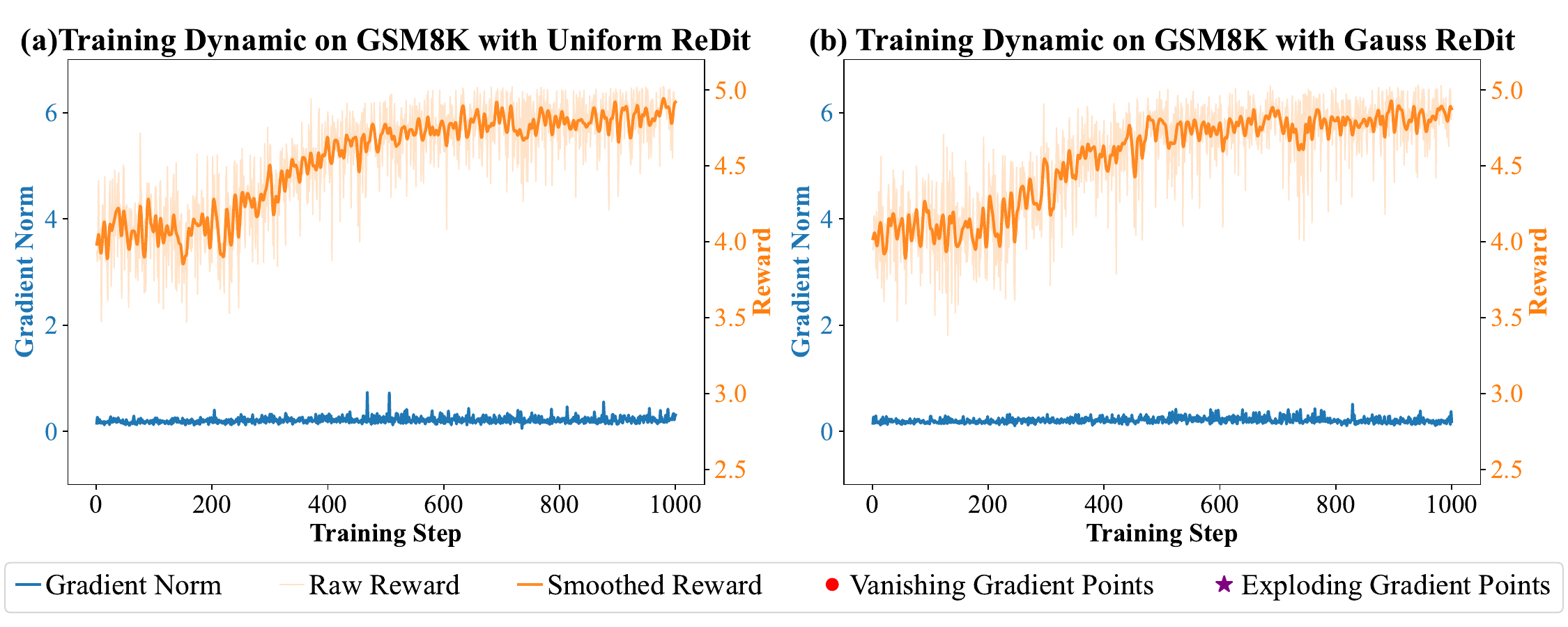}
\vspace{-5mm}
\caption{Training dynamics of gradient norm and reward on the GSM8K dataset, showing the impact of perturbations of different distributions.}
\label{fig:Training Dynamics_2}
\vspace{-3mm}
\end{figure*}

\begin{figure*}[ht]
\centering
\includegraphics[width=1.0\textwidth]{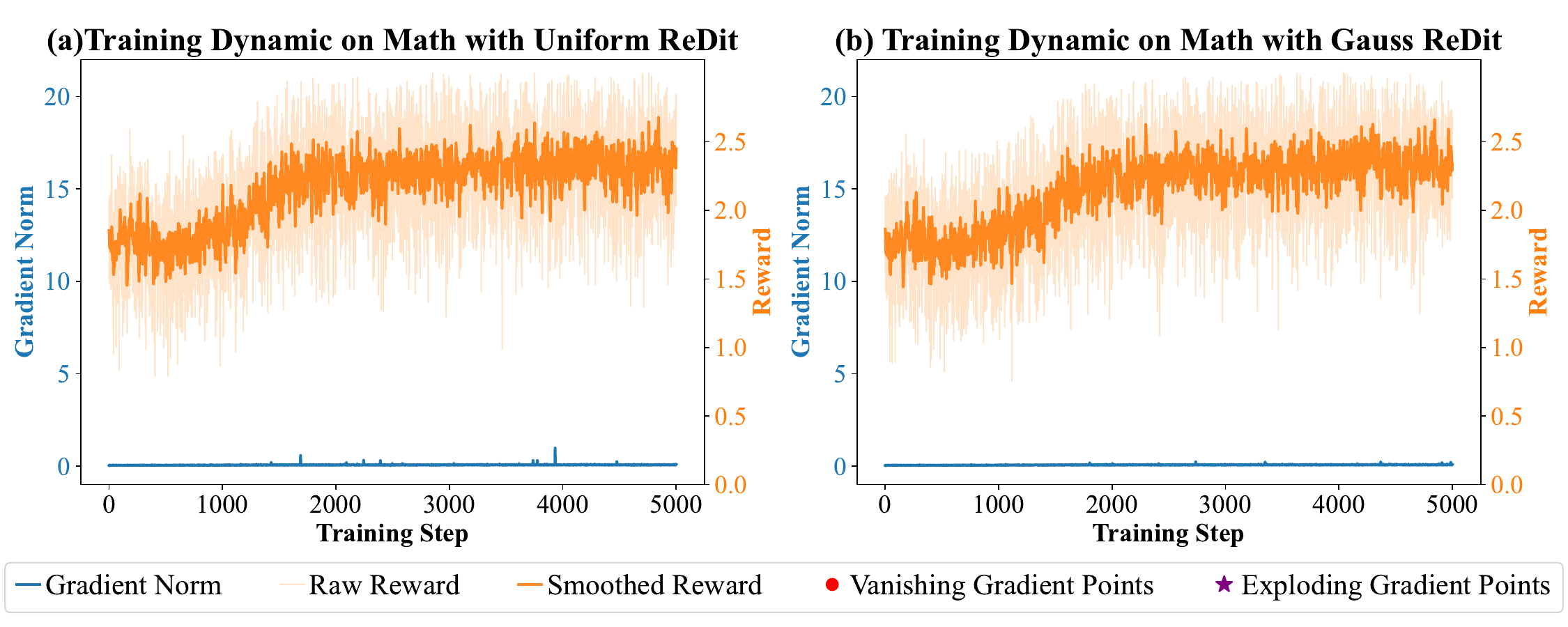}
\vspace{-5mm}
\caption{Training dynamics of gradient norm and reward on the Math dataset, showing the impact of perturbations of different distributions.}
\label{fig:Training Dynamics_3}
\vspace{-3mm}
\end{figure*}
Fig~\ref{fig:Training Dynamics_2} and Fig~\ref{fig:Training Dynamics_3}
Training dynamics using uniform and Gaussian perturbations. For both uniform and Gaussian perturbations, \ours{} shows amazing gradient stability and training stability.

\section{Experimental setting}
\label{sec:setting}
\subsection{Dataset}
\label{sec:dataset}

In this section, we introduce the statistics of the dataset and the additional processing performed on the dataset. The statistics of the dataset are shown in Table~\ref{tab:Statistics For Data}. 
\begin{table*}[ht]
\centering
\caption{Number of samples in the train, validation, and test datasets for various dateset.}
\label{tab:Statistics For Data}
\begin{tabular}{l*{4}{c}}
\toprule
\textbf{Number of samples} &\textbf{train dataset} & \textbf{validation dataset} & \textbf{test dataset} \\ 
\midrule
GSM8K & 7473 & - & 1319  \\
MATH & 7506 & - & 5003  \\
Geometry3K & 2100 & 300 & 601  \\
\bottomrule
\end{tabular}
\end{table*}

In addition, We added new templates to the original dataset to ensure the model could complete the required tasks and output formats. It is important to note that the added templates did not alter the original dataset, and special processing was performed for different LLMs. The specific examples are as follows:

\begin{tcolorbox}[title={\textbf{\small Dataset Format of GSM8K}},
colback=whitesmoke, colframe=darkblue, boxrule=2pt, arc=0mm]
{\scriptsize
\begin{lstlisting}[style=mystyle]
dataset: GSM8K
    "prompt": [
        {"role": "system", "content": "Respond in the following format: 
        <reasoning> ... </reasoning> <answer> ...</answer>"},
        {"role": "user", "content": "What is the largest single-digit prime number?"},
        {"role": "assistant", "content": "<reasoning> 9 is divisble by 3 and 8 
        is divisible by 2, but 7 is prime. </reasoning>
        <answer>7</answer>",
        {"role": "user", "content": {question}}
        ],
    "answer": {answer}
\end{lstlisting}
}
\end{tcolorbox}
\begin{tcolorbox}[title={\textbf{\small Dataset Format of MATH}},
colback=whitesmoke, colframe=darkblue, , boxrule=2pt, arc=0mm]
{\scriptsize
\begin{lstlisting}[style=mystyle]
dataset: MATH
    "prompt": [
        {"role": "system", "content": "Respond in the following format: 
        <reasoning> ... </reasoning> <answer> ...</answer>"},
        {"role": "user", "content": "{question}  
        Let"s think step by step and output the final answer within \\boxed{}."
        ],
    "answer": {answer}
\end{lstlisting}
}
\end{tcolorbox}
\begin{tcolorbox}[title={\textbf{\small Dataset Format of Geometry3K}},
colback=whitesmoke, colframe=darkblue, boxrule=2pt, arc=0mm]
{\scriptsize
\begin{lstlisting}[style=mystyle]
dataset: Geometry3K
    "prompt": [
        {"role": "user", "content": [{   
            "type": "image",
            "image": {image}, 
        },
        {
            "type": "text",
            "text": {question} + ". 
            You FIRST think about the reasoning process as an internal monologue and
            then provide the final answer. The reasoning process MUST BE enclosed 
            within <think> </think> tags. The final answer MUST BE put in \\boxed{}."
            },],
        }
    ]
    "answer": {answer}
\end{lstlisting}
}
\end{tcolorbox}

\subsection{Reward function}
\label{sec:reward}
We design five reward functions for the GSM8K dataset and show how to implement \ours{}:

\begin{tcolorbox}[title={\textbf{\small GSM8K Accuracy Reward Function}},
colback=whitesmoke, colframe=darksalmon, boxrule=2pt, arc=0mm]
{\scriptsize
\begin{lstlisting}[style=newstyle]
def correctness_reward_func_with_noise(prompts, completions, answer, **kwargs) -> list[float]:
    def extract_number(s: str) -> str:
        match = re.search(r'\d+', s)
        return match.group(0) if match else ''
    responses = [completion[0]['content'] for completion in completions]
    q = prompts[0][-1]['content']
    extracted_responses = [extract_xml_answer(r) for r in responses]
    original_rewards = [2.0 if extract_number(r) == extract_number(a) else 0.0 for r, a in zip(extracted_responses, answer)]
    
    # ReDit add
    noisy_rewards = [r + random.uniform(-m * 2.0, m * 2.0) for r in original_rewards] 
    #noisy_rewards = [r + random.gauss(0, 2.0 * m / (3 ** 0.5)) for r in original_rewards] 
    return noisy_rewards
\end{lstlisting}
}
\end{tcolorbox}

\begin{tcolorbox}[title={\textbf{\small GSM8K Int Reward Function}},
colback=whitesmoke, colframe=darksalmon, boxrule=2pt, arc=0mm]
{\scriptsize
\begin{lstlisting}[style=newstyle]
def int_reward_func_with_noise(completions, **kwargs) -> list[float]:
    responses = [completion[0]['content'] for completion in completions]
    extracted_responses = [extract_xml_answer(r) for r in responses]
    original_rewards = [0.5 if r.isdigit() else 0.0 for r in extracted_responses]
    
    # ReDit add
    noisy_rewards = [r + random.uniform(-m * 0.5, m * 0.5) for r in original_rewards]
    #noisy_rewards = [r + random.gauss(0, 0.5 * m / (3 ** 0.5)) for r in original_rewards]
    return noisy_rewards
\end{lstlisting}
}
\end{tcolorbox}

\begin{tcolorbox}[title={\textbf{\small GSM8K Strict Format Reward Function}},
colback=whitesmoke, colframe=darksalmon, boxrule=2pt, arc=0mm]
{\scriptsize
\begin{lstlisting}[style=newstyle]
def strict_format_reward_func_with_noise(completions, **kwargs) -> list[float]:
    pattern = r"^<reasoning>\n[\s\S]*?\n</reasoning>\n<answer>\n[\s\S]*?</answer>$"
    completion_contents = [completion[0]["content"].strip() for completion in completions]
    matches = [re.match(pattern, content, re.DOTALL | re.MULTILINE) for content in completion_contents]
    original_rewards = [1.0 if match else 0.0 for match in matches]
    
    # ReDit add
    noisy_rewards = [r + random.uniform(-m * 1.0, m * 1.0) for r in original_rewards]
    #noisy_rewards = [r + random.gauss(0, 1.0 * m / (3 ** 0.5)) for r in original_rewards]
    return noisy_rewards
\end{lstlisting}
}
\end{tcolorbox}

\begin{tcolorbox}[title={\textbf{\small GSM8K Sort Format Reward Function}},
colback=whitesmoke, colframe=darksalmon, boxrule=2pt, arc=0mm]
{\scriptsize
\begin{lstlisting}[style=newstyle]
def soft_format_reward_func_with_noise(completions, **kwargs) -> list[float]:
    pattern = r"^<reasoning>[\s\S]*?</reasoning>[\s\S]*?<answer>[\s\S]*?</answer>$"
    completion_contents = [completion[0]["content"].strip() for completion in completions]
    matches = [re.match(pattern, content, re.DOTALL | re.MULTILINE) for content in completion_contents]
    original_rewards = [1.0 if match else 0.0 for match in matches]
    
    #  ReDit add
    noisy_rewards = [r + random.uniform(-m * 1.0, m * 1.0) for r in original_rewards]
    #noisy_rewards = [r + random.gauss(0, 1.0 * m / (3 ** 0.5)) for r in original_rewards]
    return noisy_rewards
\end{lstlisting}
}
\end{tcolorbox}

\begin{tcolorbox}[title={\textbf{\small GSM8K Reasoning Format Reward Function}},
colback=whitesmoke, colframe=darksalmon, boxrule=2pt, arc=0mm]
{\scriptsize
\begin{lstlisting}[style=newstyle]
def xmlcount_reward_func_with_noise(completions, **kwargs) -> list[float]:
    def count_xml(text) -> float:
        count = 0.0
        if text.count("<reasoning>\n") == 1:
            count += 0.125
        if text.count("\n</reasoning>\n") == 1:
            count += 0.125
        if text.count("\n<answer>\n") == 1:
            count += 0.125
            #count -= len(text.split("\n</answer>\n")[-1])*0.001
        if text.count("\n</answer>") == 1:
            count += 0.125
            count -= (len(text.split("\n</answer>")[-1]) - 1)*0.001
        return count
    contents = [completion[0]["content"] for completion in completions]
    original_rewards = [count_xml(c) for c in contents]
    
    # ReDit add
    noisy_rewards = [r + random.uniform(-m * 0.5, m * 0.5) for r in original_rewards]
    #noisy_rewards = [r + random.gauss(0, 0.5 * m / (3 ** 0.5))  for r in original_rewards]
    return noisy_rewards
\end{lstlisting}
}
\end{tcolorbox}
As shown in the above code block, \ours{} does not need to be modified in a complex way, only the reward function needs to be modified, and any method can be easily integrated. The reward functions of other datasets can be found in the code.

\subsection{Specific experimental parameters} 
\label{sec:parameters}
In this section, we present the experimental parameters, including LoRA parameters, GRPO and other baseline experimental parameters.

\begin{table*}[ht]
\centering
\caption{LoRA Parameters}
\label{tab:lora_parameters}
\begin{tabular}{l*{4}{c}}
\toprule
\textbf{LoRA Target} & \textbf{LoRA Rank} & \textbf{LoRA Alpha} & \textbf{LoRA Dropout}  \\ 
\midrule
q \& v Proj & 8 & 64 & 0.05  \\
\bottomrule
\end{tabular}
\end{table*}

\begin{table*}[ht]
\centering
\caption{GRPO Parameters}
\label{tab:grpo_parameters}
\begin{tabular}{l*{4}{c}}
\toprule
\textbf{Learning Rate} & \textbf{Num Generations} & \textbf{Epochs}  \\ 
\midrule
5e-6 & 4 & 10   \\
\bottomrule
\end{tabular}
\end{table*}

\begin{table*}[ht]
\centering
\caption{DAPO Parameters}
\label{tab:dapo_parameters}
\begin{tabular}{l*{4}{c}}
\toprule
\textbf{Clip Ratio Low} & \textbf{Clip Ratio Low} & \textbf{Clip Ratio C} & \textbf{Num Generations Max} \\ 
\midrule
0.2  & 0.28 & 10.0 & 10  \\
\bottomrule
\end{tabular}
\end{table*}

\section{Additional experiments}
\label{additional_experiment}

\subsection{Results on the Code Generation Datasets}
\label{code}

To validate the general applicability of our method (\ours{}) beyond mathematical reasoning, we conducted a comprehensive evaluation on the domain of code generation. We performed experiments on three widely-used coding benchmarks: APPS, HumanEval, and CodeContests. This evaluation was designed to test the hypothesis that the core benefit of ReDit---stabilizing the learning signal to improve optimization---is a general principle that is not limited to a single domain.

\begin{figure*}[t]
\centering
\includegraphics[width=1.0\textwidth]{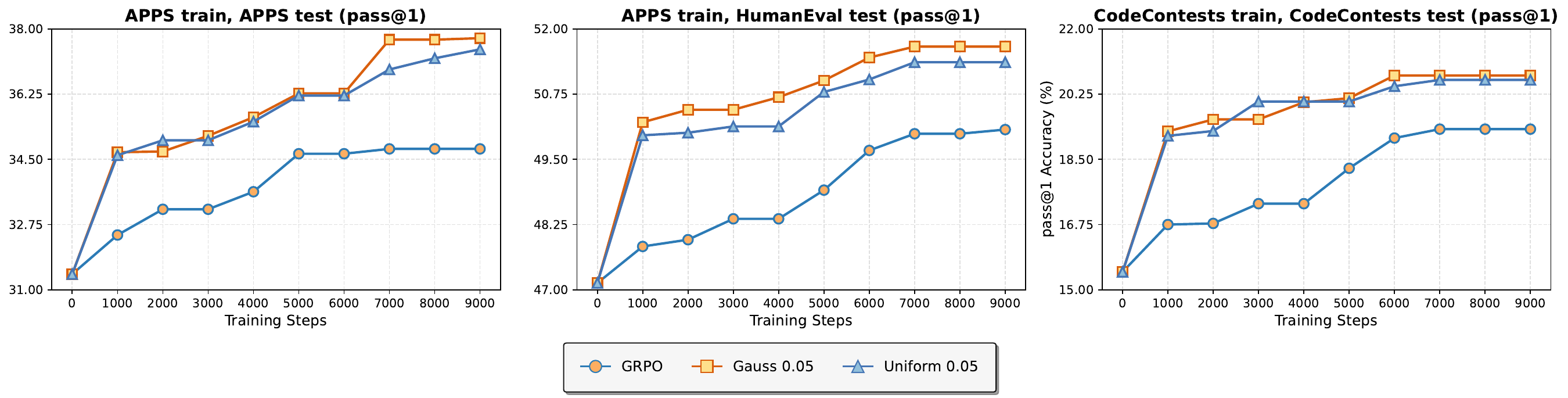}
\caption{Performance comparison on three code generation benchmarks: (left) APPS test, (center) HumanEval test, and (right) CodeContests test. The $pass@1$ accuracy is reported across training steps. Both ReDit variants (Gauss 0.05 and Uniform 0.05) consistently and significantly outperform the GRPO baseline, confirming the general applicability of our method to the coding domain.}
\label{fig:result-code}
\end{figure*}

The results are presented in \ref{fig:result-code}. The plots consistently demonstrate that both the Gaussian (Gauss 0.05) and Uniform (Uniform 0.05) variants of ReDit significantly and consistently outperform the GRPO baseline across all three coding benchmarks. On all datasets, our method not only achieves a higher final $pass@1$ accuracy but also exhibits a faster convergence rate. This strongly suggests that ReDit provides a more robust optimization pathway, and its benefits generalize effectively to complex tasks such as code generation.

\subsection{Results on Full Parameter Fine-Tuning}
\label{full}

To confirm that the benefits of ReDit are not limited to parameter-efficient fine-tuning (PEFT) methods like LoRA, we conducted additional experiments using a full parameter fine-tuning approach. This evaluation addresses whether ReDit's effectiveness is a general property of the optimization process itself, rather than an artifact of a specific tuning method \cite{wei2025paftpromptagnosticfinetuning}.

For these experiments, we utilized a setup that differs from our primary PEFT experiments; specifically, we employed the VERL framework for training with 8 GPUs. We evaluated this full fine-tuning setup on our three mathematical reasoning benchmarks: GSM8K, MATH, and Geo3k.

\begin{figure*}[t]
\centering
\includegraphics[width=1.0\textwidth]{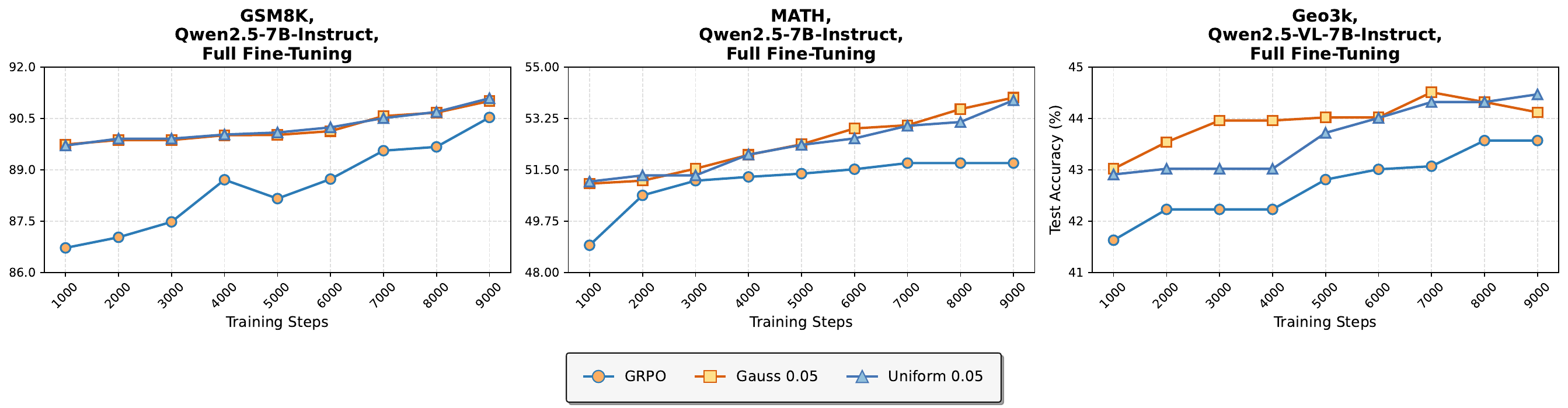}
\caption{Performance comparison on mathematical reasoning benchmarks using \textbf{full parameter fine-tuning}. The plots show test accuracy across training steps for (left) GSM8K, (center) MATH, and (right) Geo3k. In this full-tuning setting, both ReDit variants (Gauss 0.05 and Uniform 0.05) consistently achieve higher test accuracy than the GRPO baseline, confirming that our method's benefits generalize beyond parameter-efficient tuning (PEFT).}
\label{fig:result-full}
\end{figure*}

The results are presented in \ref{fig:result-full}. The plots clearly demonstrate that ReDit (both Gauss 0.05 and Uniform 0.05 variants) consistently outperforms the GRPO baseline across all three benchmarks in this demanding full-tuning setting. The performance gap is particularly notable on the MATH and Geo3k datasets, where the GRPO baseline shows signs of stagnation, while our method continues to improve. These findings confirm that ReDit is a robust and general-purpose technique, delivering consistent performance gains in both parameter-efficient and full fine-tuning paradigms.

\subsection{Results on DeepSeek Distillation Models}
\label{distillation}

A potential concern regarding our primary results is that they are predominantly focused on the Qwen2.5 model family. To further demonstrate the robustness and architectural generalizability of ReDit, we conducted additional experiments to validate its effectiveness on specialized reasoning models.

\begin{figure*}[t]
\centering
\includegraphics[width=1.0\textwidth]{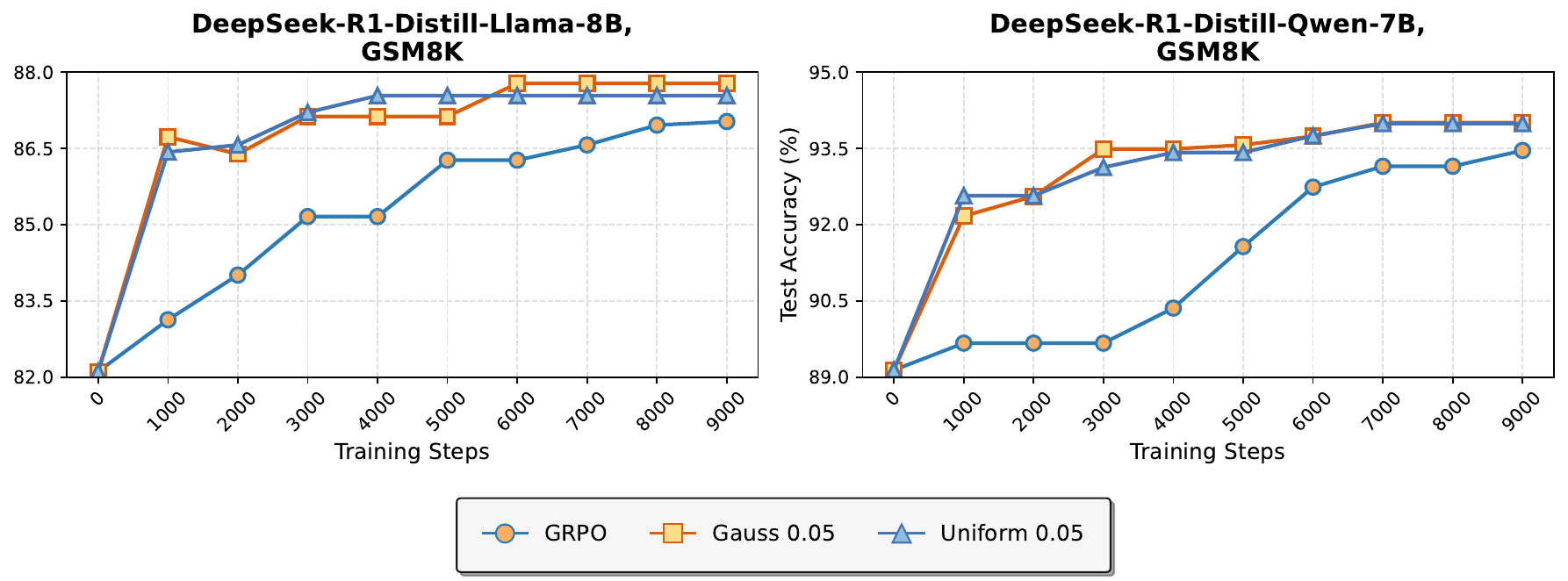}
\caption{Evaluating ReDit's generalizability on specialized reasoning models. The plots show Test Accuracy (\%) on the GSM8K benchmark for (left) \textbf{DeepSeek-R1-Distill-Llama-8B} and (right) \textbf{DeepSeek-R1-Distill-Qwen-7B}. These results confirm that ReDit's performance advantage over the GRPO baseline holds across different model architectures, not just the Qwen models used in the main experiments.}
\label{fig:result-distillation}
\end{figure*}

Specifically, we evaluated our method on two expert distillation models: \textbf{DeepSeek-R1-Distill-Llama-8B} and \textbf{DeepSeek-R1-Distill-Qwen-7B}. We used the GSM8K benchmark to compare their mathematical reasoning performance against the GRPO baseline. The results are presented in \ref{fig:result-distillation}. The plots clearly show that both the Gaussian (Gauss 0.05) and Uniform (Uniform 0.05) variants of ReDit consistently outperform the GRPO baseline on both DeepSeek models.

This finding is significant as it confirms that ReDit's ability to stabilize the training signal and improve performance is a general principle. It is not limited to a single model family but holds true across various model architectures, including those specifically optimized for reasoning tasks.

\section{More result}
\label{more_result}
In this section, we present detailed numerical results for all experiments.
\subsection{Main Result}
In this section, we show the results in Figure~\ref{fig:result-1}, the performance of GRPO and GRPO+\ours{} on different datasets.
\begin{table}[ht]
    \centering
    \small
    \setlength{\tabcolsep}{6pt}
    \caption{Performance Comparison of Different Training Steps on the Math Dataset}
    \label{tab:performance_comparison_math}
    \begin{tabular}{l *{10}{c}}
        \toprule
        Method \textbackslash Step & 0  & 1000  & 2000  & 3000  & 4000  & 5000  & 6000  & 7000  & 8000  & 9000  \\
        \midrule
        Instruct model & 39 & - & - & - & - & - & - & - & - & - \\
        GRPO & - & 47.86 & 49.46 & 47.18 & 47.28 & 47.26 & 47.57 & 47.63 & 47.89 & 48.01 \\
        Uniform \ours{}  & - & 50.02 & 50.23 & 50.34 & 50.78 & 50.96 & 51.27 & 51.37 & 51.37 & 51.96 \\
        Gauss \ours{} & - & 49.78 & 50.73 & 51.03 & 51.07 & 51.53 & 51.43 & 52.01 & 52.01 & 52.55 \\
        \bottomrule
    \end{tabular}
\end{table}

\begin{table}[ht]
    \centering
    \small
    \setlength{\tabcolsep}{6pt}
    \caption{Performance Comparison of Different Training Steps on the GSM8K Dataset}
    \label{tab:performance_comparison_gsm8k}
    \begin{tabular}{l *{10}{c}}
        \toprule
        Method \textbackslash Step & 0  & 1000  & 2000  & 3000  & 4000  & 5000  & 6000  & 7000  & 8000  & 9000  \\
        \midrule
        Instruct model & 84.91 & - & - & - & - & - & - & - & - & - \\
        GRPO & - & 85.70 & 86.01 & 86.47 & 86.73 & 87.13 & 87.78 & 88.52 & 88.73 & 89.07 \\
        Uniform \ours{}  & - & 89.16 & 89.16 & 89.31 & 89.31 & 89.31 & 89.99 & 89.99 & 89.99 & 90.76 \\
        Gauss \ours{} & - & 89.02 & 89.37 & 89.61 & 89.54 & 89.54 & 89.54 & 89.61 & 89.61 & 90.46 \\
        \bottomrule
    \end{tabular}
\end{table}

\begin{table}[ht]
    \centering
    \small
    \setlength{\tabcolsep}{6pt}
    \caption{Performance Comparison of Different Training Steps on the Geometry3K Dataset}
    \label{tab:performance_comparison_Geo}
    \begin{tabular}{l *{10}{c}}
        \toprule
        Method \textbackslash Step & 0  & 1000  & 2000  & 3000  & 4000  & 5000  & 6000  & 7000  & 8000  & 9000  \\
        \midrule
        Instruct model & 40.43 & - & - & - & - & - & - & - & - & - \\
        GRPO & - & 40.60 & 42.93 & 38.77 & 39.77 & 38.94 & 39.10 & 40.10 & 41.36 & 43.10 \\
        Uniform \ours{}  & - & 43.37 & 43.89 & 44.01 & 44.23 & 44.23 & 44.23 & 44.12 & 44.36 & 44.36 \\
        Gauss \ours{} & - & 43.67 & 43.98 & 44.03 & 44.25 & 44.25 & 44.25 & 44.25 & 44.67 & 44.67 \\
        \bottomrule
    \end{tabular}
\end{table}
Tables ~\ref{tab:performance_comparison_math}, ~\ref{tab:performance_comparison_gsm8k},~\ref{tab:performance_comparison_Geo} show the comparison of \ours{} on different datasets. \ours{} significantly improves the convergence speed of GRPO. At any same step, \ours{} achieves better performance.

\subsection{Baseline Result}
In this section, we present all numerical results in Fig.~\ref{fig:result-2}.
\begin{table}[ht]
    \centering
    \small
    \setlength{\tabcolsep}{7pt}
    \caption{Performance Comparison at Different Training Steps on Different Baseline}
    \label{tab:performance_comparison_baseline}
    \begin{tabular}{l *{10}{c}}
        \toprule
        Method \textbackslash Step & 1000  & 2000  & 3000  & 4000  & 5000  & 6000  & 7000  & 8000  & 9000  \\
        \midrule
        DAPO  & 84.99 & 86.20 & 86.35 & 86.35& 86.75 & 87.04 & 87.12 & 87.17 & 87.52 \\
        Uniform \ours{}  & 87.03 & 87.15 & 87.26 & 87.54 & 87.54 & 87.69 & 87.83 & 88.03 & 88.57 \\
        Gauss \ours{}  & 87.76 & 87.96 & 88.01 & 88.01 & 88.10 & 88.37 & 88.67 & 88.96 & 89.34 \\
        \midrule
        DR.GRPO  & 84.69 & 84.23 & 84.53 & 84.91 & 85.67 & 85.67 & 85.67 & 85.90 & 86.13 \\
        Uniform \ours{}  & 86.27 & 86.36 & 86.45 & 86.54 & 86.75 & 87.03 & 87.26 & 87.16 & 87.34 \\
        Gauss \ours{}  & 86.47 & 86.23 & 87.10 & 87.16 & 87.56 & 87.67 & 87.67 & 87.67 & 87.69 \\
        \midrule
        REINFORCE++  & 84.91 & 84.69 & 85.06 & 85.14 & 85.14 & 85.14 & 86.10 & 86.17 & 86.25 \\
        Uniform \ours{}  & 86.21 & 86.11 & 86.67 & 86.31 & 86.75& 87.01 & 87.26 & 87.59 & 87.59 \\
        Gauss \ours{}  & 86.17 & 86.27 & 86.47 & 86.83 & 86.83 & 87.06 & 87.63 & 87.76 & 87.96 \\
        \bottomrule
    \end{tabular}
\end{table}
As shown in Table~\ref{tab:performance_comparison_baseline}, we demonstrate the effect of using \ours{} on GSM8K based on the GRPO improvement method. The experimental results show that \ours{} can also improve the convergence speed and performance on these algorithms.
\subsection{Variance  Result}
\begin{figure*}[ht]
\centering
\includegraphics[width=0.7\textwidth]{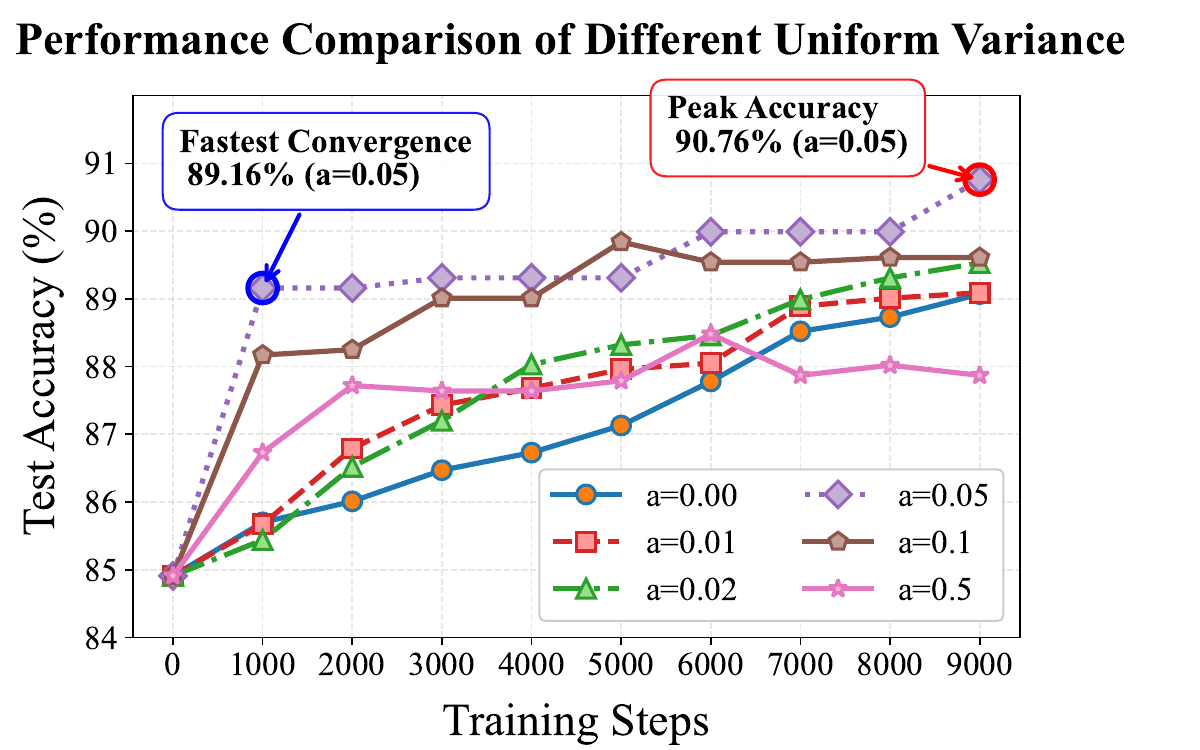}
\caption{\ours{} uniform perturbation performance changes with variance.}
\label{fig:uniform_variance}
\end{figure*}

In this section, we show more results on the performance of \ours{} as the perturbation changes. As shown in Figure ~\ref{fig:uniform_variance}, the variance of uniform perturbation is similar to the variance of Gaussian perturbation, and the appropriate variance can achieve the best performance. The specific numerical results are shown in Tables ~\ref{tab:gauss_performance_comparison} and ~\ref{tab:uniform_performance_comparison}.

\begin{table}[ht]
    \centering
    \small
    \setlength{\tabcolsep}{7pt}
    \caption{Performance Comparison of Different variance on the Gauss Perturbation}
    \label{tab:gauss_performance_comparison}
    \begin{tabular}{l *{9}{c}}
        \toprule
        Variance \textbackslash Step & 1000  & 2000  & 3000  & 4000  & 5000  & 6000  & 7000  & 8000  & 9000  \\
        \midrule
        0.01 & 85.97 & 87.01 & 87.40 & 87.54 & 87.92 & 88.76 & 88.84 & 89.54 & 89.54 \\
        0.02 & 86.40 & 87.70 & 88.16 & 89.23 & 89.39 & 90.22 & 90.14 & 90.14 & 90.14 \\
        0.05 & 89.02 & 89.37 & 89.61 & 89.54 & 89.54 & 89.54 & 89.61 & 89.61 & 90.46 \\
        0.1  & 87.64 & 89.08 & 89.69 & 89.84 & 90.07 & 89.84 & 89.84 & 89.84 & 90.07 \\
        0.3  & 87.87 & 88.48 & 88.78 & 88.93 & 89.39 & 89.39 & 89.39 & 89.46 & 89.46 \\
        0.5 & 86.81 & 87.57 & 87.41 & 87.64 & 87.64 & 87.95 & 88.32 & 88.48 & 88.95 \\
        \bottomrule
    \end{tabular}
\end{table}

\begin{table}[ht]
    \centering
    \small
    \setlength{\tabcolsep}{7pt}
    \caption{Performance Comparison of Different variance on the Uniform Perturbation}
    \label{tab:uniform_performance_comparison}
    \begin{tabular}{l *{9}{c}}
        \toprule
        Variance \textbackslash Step & 1000  & 2000  & 3000  & 4000  & 5000  & 6000  & 7000  & 8000  & 9000  \\
        \midrule
        0.01 & 85.67 & 86.79 & 87.43 & 87.68 & 87.96 & 88.05 & 88.89 & 89.01 & 89.09 \\
        0.02 & 85.44 & 86.52 & 87.20 & 88.03 & 88.32 & 88.46 & 88.99 & 89.31 & 89.53 \\
        0.05 & 89.16 & 89.16 & 89.31 & 89.31 & 89.31 & 89.99 & 89.99 & 89.99 & 90.76 \\
        0.1 & 88.17 & 88.25 & 89.01 & 89.01 & 89.84 & 89.54 & 89.54 & 89.61 & 89.61 \\
        0.3 & 87.49 & 88.25 & 88.25 & 88.02 & 88.17 & 87.95 & 88.93 & 88.70 & 88.78 \\
        0.5 & 86.73 & 87.72 & 87.64 & 87.64 & 87.79 & 88.48 & 87.87 & 88.02 & 87.87 \\
        \bottomrule
    \end{tabular}
\end{table}

\subsection{Scheduled Perturbation Result}
\label{sec: scheduled}
\begin{figure*}[ht]
\centering
\includegraphics[width=1.0\textwidth]{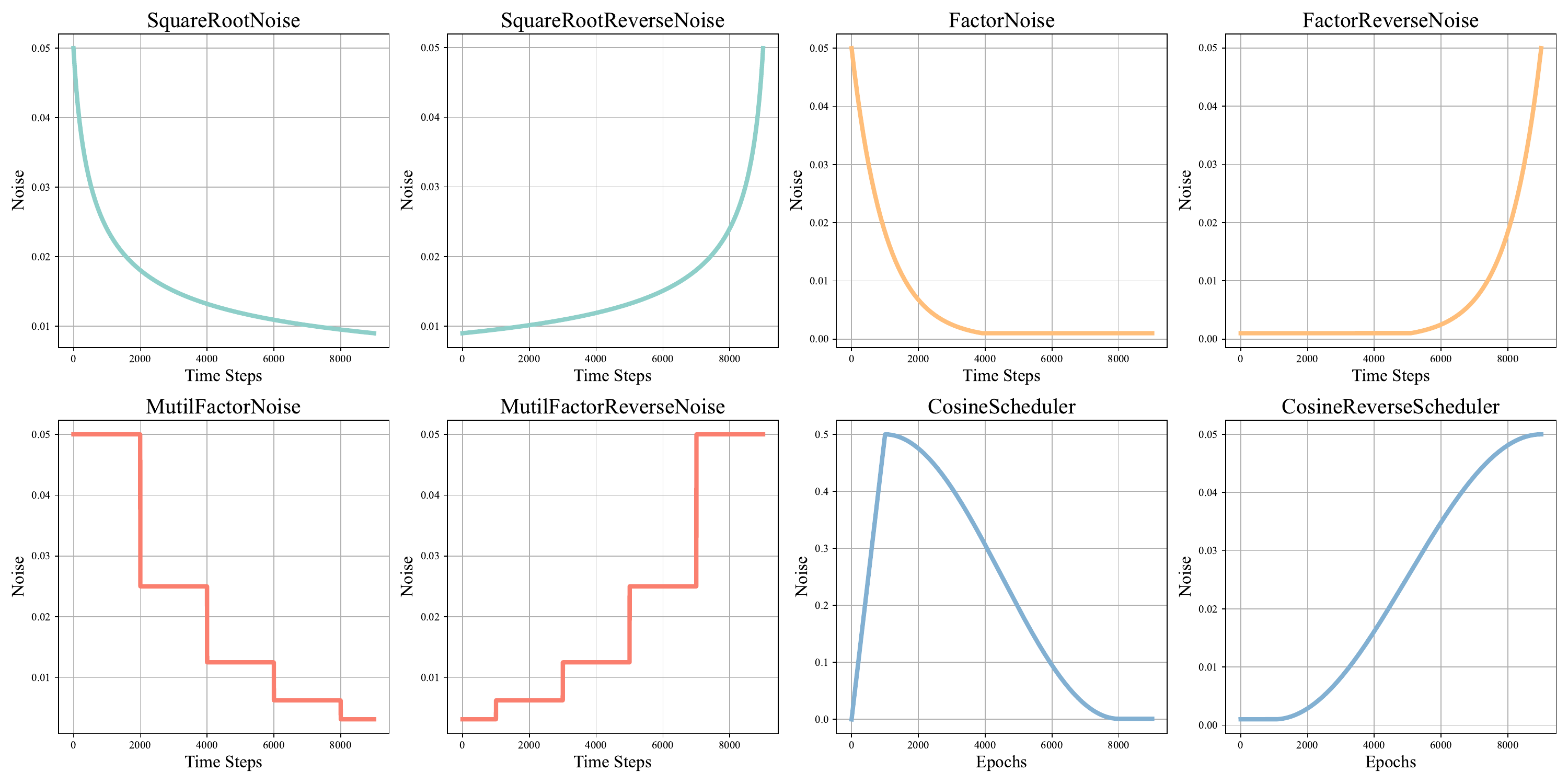}
\caption{\ours{} scheduled perturbation Variance trend with training step (taking the original variance as 0.05 as an example)}
\label{fig:schedule}
\end{figure*}

In this section, we show the changing trends of different scheduled perturbation strategies, as shown in Figure~\ref{fig:schedule}. We took the perturbation of Gauss distribution as an example and conducted experiments. The experimental results are shown in Table~\ref{tab:perturbation_methods}. The CosineReverse strategy shows the best performance.

\begin{table}[ht]
    \centering
    \small
    \setlength{\tabcolsep}{7pt}
    \caption{Performance Comparison of Different Scheduled Perturbation Methods}
    \label{tab:perturbation_methods}
    \begin{tabular}{l *{9}{c}}
        \toprule
        Method \textbackslash Step & 1000 & 2000  & 3000  & 4000  & 5000  & 6000  & 7000  & 8000  & 9000  \\
        \midrule
        SquareRoot   & 88.10 & 89.31 & 88.93 & 89.69 & 89.46 & 89.46 & 89.46 & 89.46 & 90.22 \\
        SquareRootReverse   & 88.55 & 89.54 & 89.46 & 90.07 & 90.07 & 89.31 & 89.61 & 89.54 & 89.69 \\
        Factor   & 88.25 & 88.63 & 89.69 & 89.46 & 89.23 & 89.54 & 89.46 & 89.31 & 89.69 \\
        FactorReverse   & 88.48 & 88.32 & 89.39 & 88.78 & 88.93 & 89.54 & 89.61 & 89.76 & 89.46 \\
        MutilFactor   & 87.87 & 89.31 & 89.01 & 89.01 & 89.01 & 89.61 & 89.16 & 89.61 & 89.46 \\
        MutilFactorReverse   & 88.17 & 88.78 & 88.86 & 89.01 & 88.93 & 88.93 & 89.39 & 89.16 & 89.54 \\
        Cosine   & 88.32 & 88.32 & 89.39 & 89.84 & 89.76 & 89.61 & 90.14 & 90.46 & 90.23 \\
        CosineReverse   & 89.08 & 87.95 & 89.54 & 89.08 & 89.16 & 90.37 & 90.07 & 90.84 & 91.84 \\
        \bottomrule
    \end{tabular}
\end{table}

\end{document}